\documentclass{article} 
\PassOptionsToPackage{dvipsnames}{xcolor}
\usepackage{iclr2025_conference,times}

\usepackage{subcaption}
\usepackage{booktabs}
\usepackage{tabularx}
\usepackage{float}
\usepackage{graphicx}
\usepackage{caption}
\usepackage[margin=1in]{geometry} 

\usepackage{listings}
\definecolor{codegreen}{rgb}{0,0.6,0}
\definecolor{codegray}{rgb}{0.5,0.5,0.5}
\definecolor{codepurple}{rgb}{0.58,0,0.82}
\definecolor{backcolour}{rgb}{0.95,0.95,0.92}
\lstdefinestyle{mystyle}{
    backgroundcolor=\color{backcolour},   
    commentstyle=\color{codegreen},
    keywordstyle=\color{magenta},
    numberstyle=\tiny\color{codegray},
    stringstyle=\color{codepurple},
    basicstyle=\ttfamily\footnotesize,
    breaklines=true, 
    breakindent=0pt,
    keepspaces=true,                 
    numbersep=5pt,                  
    showspaces=false,                
    showstringspaces=false,
    frame=single,
    columns=fullflexible,  
    literate=
    {•}{{*}}1
    {–}{{-}}1
    {—}{{-}}1
    {‐}{{-}}1
    {-}{{-}}1
    {“}{{"}}1
    {”}{{"}}1
    {‘}{{'}}1
    {’}{{'}}1
    {…}{{...}}1
    {⇒}{{$\Rightarrow$}}1
    {“}{{"}}1
    {”}{{"}}1
    {≤}{{$\leq$}}1
    {→}{{$\rightarrow$}}1
}
\usepackage{amsmath,amssymb,amsthm,mathtools,bm,bbm}
\usepackage{enumitem}
\usepackage{hyperref}
\usepackage{caption}
\usepackage{array}
\newcolumntype{C}[1]{>{\centering\arraybackslash}m{#1}}
\usepackage{pgfplots}
\pgfplotsset{compat=1.18}
\usepackage{subcaption}
\usepackage{wrapfig}

\newcommand{\defeq}{\vcentcolon=}

\newcommand{\Conf}{\mathcal{C}}
\newcommand{\GraphFam}{\mathfrak{G}}    
\newcommand{\Ebb}{\mathbb{E}}
\newcommand{\Dist}{\mathsf{Dist}}       
\newcommand{\1}{\mathbf{1}}

\definecolor{mybarcolor}{RGB}{77,171,247}
\newcommand{\barcolor}{mybarcolor}
\newcommand{\baselineOpacity}{0.5}
\newcommand{\hlcolor}{mybarcolor}

\usepackage{tikz}
\usetikzlibrary{arrows.meta,positioning,calc,shapes.geometric}

\usepackage[most]{tcolorbox}
\usepackage{enumitem}

\definecolor{promptbg}{RGB}{233,240,255}
\definecolor{annotbg}{RGB}{255,250,200}
\definecolor{highlight}{RGB}{252,213,122}

\usepackage{listings}
\usepackage[most]{tcolorbox}
\tcbuselibrary{listingsutf8}
\newtcolorbox{instructbox}[1][]{
  breakable,
  enhanced,
  colback=gray!4,
  colframe=black!50,
  sharp corners,
  fontupper=\small\ttfamily,
  left=8pt,
  right=8pt,
  top=6pt,
  bottom=6pt,
  boxrule=0.6pt,
  listing only,
  listing options={
    basicstyle=\ttfamily\small,
    breaklines=true,
    postbreak=\mbox{\textcolor{gray}{$\hookrightarrow$}\space},
    showstringspaces=false,
    tabsize=2,
    numbers=none,
    stepnumber=1,
    xleftmargin=0pt,
    xrightmargin=0pt,
    frame=none,
    keywordstyle=\color{blue}\bfseries,
    emph={curr_instructions,inputs_outputs_feedback},
    emphstyle=\color{teal}
  },
  title=#1
}
\usepackage{listings}
\usepackage{caption}
\usepackage{subcaption}
\usepackage{arydshln}
\usepackage{algorithm}
\usepackage{algpseudocode}

\usepackage{amsmath,amsfonts,bm}

\def\eqref#1{equation~\ref{#1}}

\def\1{\bm{1}}

\DeclareMathAlphabet{\mathsfit}{\encodingdefault}{\sfdefault}{m}{sl}
\SetMathAlphabet{\mathsfit}{bold}{\encodingdefault}{\sfdefault}{bx}{n}

\usepackage{hyperref}
\usepackage{url}

\title{Maestro: Joint Graph \& Config Optimization\\ for Reliable AI Agents}

\author{
Wenxiao Wang \\ 
\texttt{wwx@relai.ai}\\
RELAI.ai\\
\And Priyatham Kattakinda \\
\texttt{priyatham@relai.ai}\\
RELAI.ai\\
\And Soheil Feizi\\
\texttt{sfeizi@relai.ai}\\
RELAI.ai\\
}

\date{}

\iclrfinalcopy 
\begin{document}

\maketitle

\begin{abstract}
Building reliable LLM agents requires decisions at two levels: the \emph{graph} (which modules exist and how information flows) and the \emph{configuration} of each node (models, prompts, tools, control knobs). Most existing optimizers tune configurations while holding the graph fixed, leaving structural failure modes unaddressed. We introduce \textbf{Maestro}, a framework-agnostic holistic optimizer for LLM agents that \emph{jointly} searches over graphs and configurations to maximize agent quality, subject to explicit rollout/token budgets. Beyond numeric metrics, Maestro leverages \emph{reflective textual feedback} from traces to prioritize edits, improving sample efficiency and targeting specific failure modes. On the IFBench and HotpotQA benchmarks, Maestro consistently surpasses leading prompt optimizers—MIPROv2, GEPA, and GEPA+Merge—by an average of 12\%, 4.9\%, and 4.86\%, respectively; even when restricted to prompt-only optimization, it still leads by 9.65\%, 2.37\%, and 2.41\%. Maestro achieves these results with far fewer rollouts than GEPA. We further show large gains on two applications (interviewer \& RAG agents), highlighting that joint graph \& configuration search addresses structural failure modes  that prompt tuning alone cannot fix.

\end{abstract}

\section{Introduction}

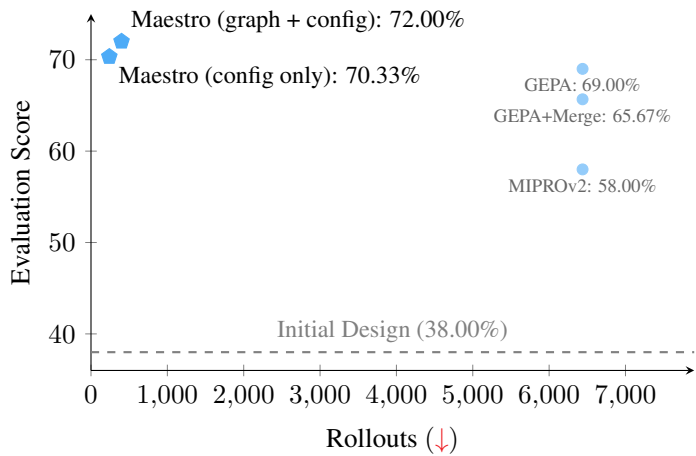
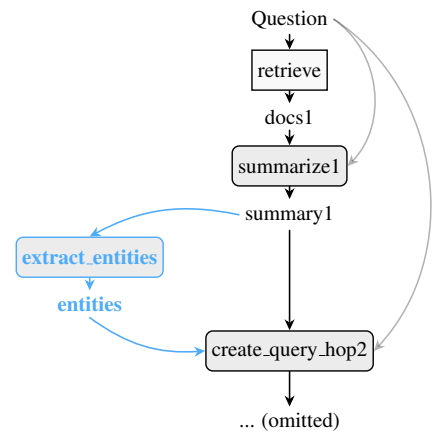
\begin{figure}[h!]
\centering

\begin{minipage}{0.95\textwidth}
\centering
\begin{subfigure}{0.6\textwidth}
\centering
\resizebox{\linewidth}{!}{%
\begin{tikzpicture}
\begin{axis}[
    width=\linewidth,       
    height=0.66\linewidth,
    xlabel={Rollouts $(\textcolor{Red}{\downarrow})$},
    ylabel={Evaluation Score},
    axis lines=left,
    ymin=36, ymax=75,      
    xmin=0, xmax=7900,      
    xticklabel style={font=\normalsize},
    every axis label/.style={font=\normalsize},        
    every node near coord/.style={font=\normalsize},   
    nodes near coords,
    point meta=explicit symbolic,                 
    clip=false,
]


\addplot[only marks, mark=*, draw=\barcolor, fill=\barcolor, opacity=0.6,
    every node near coord/.append style={
        anchor=north,   
        font=\scriptsize
    }] coordinates {
    (6438,58.00) [MIPROv2: 58.00\%]
    (6438,69.00) [GEPA: 69.00\%]
    (6438,65.67) [GEPA+Merge: 65.67\%]
};

\addplot[only marks, mark=pentagon*, draw=\barcolor, fill=\barcolor,
    every node near coord/.append style={
        anchor=south west,   
        font=\small
    }, mark options={fill=\barcolor, scale=1.5},] coordinates {
    (400,72.00) [Maestro (graph + config): 72.00\%]
};

\addplot[only marks, mark=pentagon*, draw=\barcolor, fill=\barcolor,
    every node near coord/.append style={
        anchor=north west,   
        font=\small
    }, mark options={fill=\barcolor, scale=1.5},] coordinates {
    (240,70.33) [Maestro (config only): 70.33\%]
};

\addplot[dashed, thick, gray, no marks] 
    coordinates {(0,38) (7900,38)} 
    node[pos=0.5, anchor=south, font=\footnotesize] {Initial Design (38.00\%)};


\end{axis}
\end{tikzpicture}
        }
        \caption{Evaluation Score vs. Rollouts on HotpotQA (gpt-4.1-mini).}
    \end{subfigure}
    \hfill
    \begin{subfigure}{0.38\textwidth}
        \centering
        \resizebox{\linewidth}{!}{%
\begin{tikzpicture}[
        >=Stealth,
        line cap=round,
        llm/.style={draw, rounded corners, minimum width=14mm, minimum height=8mm,
                    align=center, thick, fill=gray!15, font=\large},
        tool/.style={draw, minimum width=12mm, minimum height=8mm,
                     align=center, thick, fill=gray!5, font=\large},
        hyperparams/.style={draw, minimum width=12mm, minimum height=8mm,
                     align=center, thick, fill=gray!5, font=\large},
        database/.style={cylinder, draw, shape border rotate=90, aspect=0.25,
                         minimum height=10mm, minimum width=12mm, fill=gray!20, font=\large},
        io_text/.style={font=\large, align=center},
        edge/.style={-{Stealth[length=2.2mm,width=2mm]}, line width=0.9pt},
        double_edge/.style={<->, {Stealth[length=2.2mm,width=2mm]}, line width=0.9pt} 
    ]

    \node[io_text] (question) {Question};
    \node[tool, below=0.3cm of question] (retrieve1) {retrieve};
    \node[io_text, below=0.25cm of retrieve1] (docs1) {docs1};
    \node[llm, below=0.3cm of docs1] (summarize1) {summarize1};
    \node[io_text, below=0.25cm of summarize1] (summary1) {summary1};
    \node[llm, below=2cm of summary1] (createqueryhop2) {create\_query\_hop2};
    \node[io_text, below=0.7cm of createqueryhop2] (query2) {... (omitted)};

    \node[llm, below=0.3cm of summary1, draw=\hlcolor] (extractentities) at ($ (summary1)!0.05!(createqueryhop2) + (-4cm,0cm) $) {\textcolor{\hlcolor}{\textbf{extract\_entities}}};
    \node[io_text, below=0.25cm of extractentities] (entities) {\textcolor{\hlcolor}{\textbf{entities}}};
    \draw[edge, color=\hlcolor] (extractentities) to (entities);
    \draw[edge, bend right=20, color=\hlcolor] (summary1.west) to (extractentities.north);
    \draw[edge, bend right=20, color=\hlcolor] (entities.south) to (createqueryhop2.west);
    
    \draw[edge] (question) to (retrieve1);
    \draw[edge] (retrieve1) to (docs1);
    \draw[edge] (docs1) to (summarize1);
    \draw[edge, bend left=50, opacity=0.3] (question.east) to (summarize1.east);
    \draw[edge] (summarize1) to (summary1);
    \draw[edge] (summary1) to (createqueryhop2);
    \draw[edge, bend left=50, opacity=0.3] (question.east) to (createqueryhop2.east);
    \draw[edge] (createqueryhop2) to (query2);

    \end{tikzpicture}%
}
\caption{Agent graph edits suggested by \textbf{Maestro} for HotpotQA; \textcolor{\hlcolor}{changes highlighted}.}
\end{subfigure}
\end{minipage}
    \caption{Maestro outperforms leading prompt optimizers with far fewer rollouts.}
    \label{fig:lead}
\end{figure}

Large Language Models (LLMs) have enabled a new \emph{agentic AI} paradigm, in which AI agents autonomously plan and act to accomplish complex tasks. These LLM-based agents aim to reduce the need for human orchestration by converting high-level instructions into multi-step decisions and tool calls~\cite{ref_agentsurvey2024}. From customer support to code generation, such agents integrate capabilities like tool use, memory, and self-reflection to handle tasks beyond a single prompt’s scope~\cite{ref_agentscope2024,ref_memgpt2024}. Some even argue that the agent paradigm could be a stepping stone toward artificial general intelligence (AGI) by virtue of its autonomy and cross-domain adaptability~\cite{ref_openai2023agi}. Despite this promise, current AI agents often fall short of expectations~\cite{ref_stateofagents2024}, frequently failing to deliver reliable results in practice. This gap between potential and reality stems largely from \textbf{limitations in today’s agent designs}, which undermine robustness and generality.

Current LLM-based agents suffer from several \textbf{technical limitations} that hinder their effectiveness, scalability and reliability including:
\begin{itemize}
    \item \textbf{Limited Instruction-Following Capacity:} Agents built on smaller or less-tuned models can misunderstand or deviate from given instructions~\cite{ref_autogptfail2023}. Even large models can go off course without carefully crafted prompts or persona constraints, leading to inconsistent compliance with user intent.
    \item \textbf{Unanticipated Corner-Case Failures:} Agents are brittle when faced with scenarios outside their training or design assumptions~\cite{ref_agentscope2024}. They may perform well on the happy path but break in corner cases, yielding incoherent or incorrect behavior in novel situations.
    \item \textbf{Global State Mismanagement:} Without persistent memory, agents struggle to maintain coherent global state across steps or between sub-agents~\cite{ref_memgpt2024}. Important information can be forgotten or overwritten, especially in long-horizon tasks.
    \item \textbf{Architectural Fragility:} Complex pipelines of prompts and tools can be fragile—minor changes in one component can break the chain~\cite{ref_stateofagents2024}.
    \item \textbf{Weak Error Handling \& Recovery:} Many agents lack robust mechanisms to detect and recover from errors, leading to loops or stalls~\cite{ref_autogptfail2023}.
    \item \textbf{Context-Window \& Memory Constraints:} LLMs have finite context windows, forcing reliance on external memory tools with their own retrieval issues~\cite{ref_memgpt2024}.
    \item \textbf{Poor Modularity \& Extensibility:} Many agents are one-off solutions with tightly coupled prompts and logic, making adaptation difficult~\cite{langchain2023}.
\end{itemize}

\paragraph{Challenges of building reliable AI agents.}
Constructing reliable AI agents requires optimizing \emph{both} their \textit{structural design} (the agent graph/topology) and their \textit{operational configuration} (models, prompts, tools, memory policies, and hyperparameters) in a \emph{joint}, feedback-driven manner. This optimization is challenging: the search space is high dimensional; variables are strongly dependent; variables are of mixed types; and the objective is often multi-criteria, combining measures of agent quality ranging from task performance to latency and cost.

\paragraph{Limitations of prior work in agent optimization.}
Prior work typically addresses only a slice of this space.  
Prompt-centric methods such as GEPA and MIPROv2 search the instruction space to improve adherence and task accuracy, but they presume a fixed pipeline and cannot remedy structural deficiencies (e.g., missing tools or inadequate task decomposition)~\cite{gepa2025,miprov2}.  
RL-based approaches (e.g., GRPO) adapt model parameters given scalar rewards but require many rollouts and likewise leave the agent graph unchanged~\cite{grpo}.  
MAAS takes an important step by coupling prompt and topology search; however, its search space is restricted to a predefined set of block types with largely fixed sequential connectivity, offering no explicit mechanisms for persistent state, selective context routing, or flexible divide-and-conquer graphs~\cite{maas2025}.  
Reflection-based controllers (e.g., Reflexion) attach verbal self-critique to a predetermined architecture but do not conduct global, cross-component optimization~\cite{reflexion2023}.  
Finally, popular tooling frameworks (LangChain, Haystack, AutoGPT) provide modular components for building agents but leave the optimization burden to practitioners, who must hand-tune prompts, tool use, memory, and control flow per task~\cite{langchain2023,haystack2024,ref_autogptfail2023}.  

Taken together, these lines of work reveal three gaps:  
(i) \textbf{incomplete coverage} of the design space (structure or configuration, but rarely both);  
(ii) \textbf{rigid search spaces} that prevent discovery of richer topologies (e.g., DAGs with conditional routing, shared/persistent state, or gated visibility of intermediate artifacts); and  
(iii) \textbf{under-utilization of non-numerical signals}, where qualitative evaluator feedbackis ignored in favor of scalar scores, limiting sample efficiency and the ability to target specific failure modes.  

\paragraph{Maestro: a holistic agent optimizer.}
Holistic agent optimization tunes \emph{all facets} of an AI agent—from model selection and prompt phrasing to tool usage and memory handling—in an integrated fashion. 
A \emph{holistic} optimizer should therefore (a) operate over a flexible agent-graph space that includes branching, memory/state nodes, and tool-augmented subroutines; (b) jointly tune prompts, model choices, tools, and hyperparameters; and (c) fuse \emph{both} numeric and textual feedback to drive targeted, sample-efficient improvements.

In this technical report, we introduce a {\it beta} version of \textbf{Maestro}, a framework-agnostic, holistic agent optimizer that jointly searches the \emph{graph} and \emph{configuration} dimensions of agentic systems. Maestro treats an agent as a typed computation graph whose nodes encapsulate capabilities (LLM invocation, retrieval, tool calls, memory modules, validators) and whose edges define information flow and control logic. Its search space admits (i) non-sequential DAGs with conditional routing and retries; (ii) persistent/global state with explicit read/write policies; (iii) context gating to control which intermediate artifacts are visible to which nodes; and (iv) multi-model, multi-tool choices per node with tunable hyperparameters.

Beyond aggregate scores, Maestro ingests \emph{non-numerical reflective feedback} (evaluator rubrics, free-form critiques, etc.). These signals are distilled into actionable edits to the agent design, both graph and configurations. This enables targeted fixes for instruction drift, looping, state loss, and brittle control flow that scalar-only objectives struggle to correct.

\paragraph{Results.}
On public benchmarks, Maestro surpasses leading prompt optimizers (GEPA and MIPROv2) even when optimizing prompts alone. 
On HotpotQA~\citep{yang2018hotpotqa}, Maestro-optimized prompts achieve an evaluation score of 70.33\% with as few as 240 rollouts, outperforming the previous SOTA of 69.00\% from GEPA, which required over 6000 rollouts. 
On IFBench~\citep{ifbench}, Maestro-optimized prompts achieve an evaluation score of 56.12\% with 700 rollouts, slightly exceeding GEPA's peak performance of 55.95\% after more than 3000 rollouts (with its peak reached at 678 rollouts).
\textbf{The improvements enlarge further with Maestro-optimized agent graphs.}
On HotpotQA, Maestro-optimized graphs combined with Maestro-optimized prompts reach 72.00\% with 420 rollouts and 72.33\% with 2220 rollouts, representing gains of more than 3\% compared to GEPA and MIPROv2, both requiring over 6000 rollouts. 
On IFBench, Maestro-optimized graphs and prompts achieve 59.18\% after 900 rollouts, surpassing the peak performance of GEPA and MIPROv2 (with over 3000 rollouts) by more than 3\%.

In two application settings—an interviewer agent and an AI assistant—Maestro improves over the initial agents by an average of \mbox{~41.9\%} (64\% and 19.7\%, respectively) with configuration-only optimization, and by an average of \mbox{~65.7\%} (90\% and 41.3\%, respectively) with holistic optimization. Qualitatively, Maestro reduces instruction drift, prevents loops through revised control logic, preserves global state via explicit memory nodes, and improves corner-case reliability through context gating and validator insertion.

\noindent Collectively, these results position Maestro as a \emph{robust, generalizable} solution for end-to-end agent improvement: it closes structural and configuration gaps simultaneously, exploits rich feedback for sample-efficient search, and operates across frameworks without requiring agents to be rewritten in a specific DSL~\citet{dspy2023}.

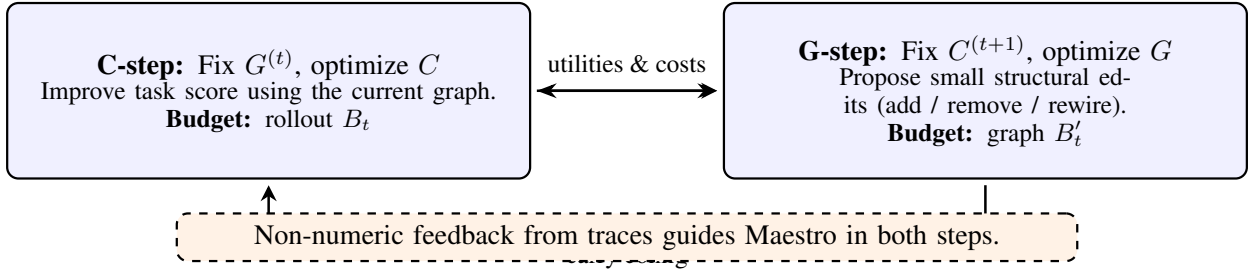
\begin{figure}[t]
\centering
\resizebox{\linewidth}{!}{%
\begin{tikzpicture}[
  >=Stealth,
  node distance=20mm and 20mm,
  box/.style={
    draw, rounded corners, thick, fill=blue!6,
    align=center, inner sep=5pt,
    minimum width=62mm, text width=62mm,  
    minimum height=22mm                    
  },
  note/.style={
    draw, rounded corners, dashed, thick, fill=orange!10,
    align=center, inner sep=4pt, text width=110mm
  },
  lab/.style={font=\footnotesize},
  edge/.style={-{Stealth[length=2mm,width=2mm]}, line width=0.9pt, shorten >=2pt, shorten <=2pt}
]
\node[box] (Cstep) {\textbf{C-step:} Fix $G^{(t)}$, optimize $C$\\[-1pt]
\footnotesize Improve task score using the current graph.\\[-1pt]
\footnotesize \textbf{Budget:} rollout $B_t$};

\node[box, right=24mm of Cstep] (Gstep) {\textbf{G-step:} Fix $C^{(t+1)}$, optimize $G$\\[-1pt]
\footnotesize Propose small structural edits (add / remove / rewire).\\[-1pt]
\footnotesize \textbf{Budget:} graph $B'_t$};

\draw[edge] (Cstep.east) -- node[lab, above, yshift=1mm]{utilities \& costs} (Gstep.west);

\draw[<->, line width=0.9pt, shorten >=2pt, shorten <=2pt] (Cstep.east) -- (Gstep.west);

\path (Cstep.south) ++(0,-7mm) coordinate (cbot);
\path (Gstep.south) ++(0,-7mm) coordinate (gbot);
\draw[edge] (Gstep.south) -- (gbot) -- node[lab, below, pos=0.5]{carry config} (cbot) -- (Cstep.south);

\node (mid) at ($(Cstep)!0.5!(Gstep)$) {};
\node[note, below=14mm of mid] (refl) {Non-numeric feedback from traces guides Maestro in both steps.};

\end{tikzpicture}%
}
\caption{\textbf{Maestro’s two complementary optimization steps.}
\emph{C-step (config update):} keep the graph fixed and tune the configuration under a rollout budget.
\emph{G-step (graph update):} keep the configuration fixed and try small graph edits under a graph budget.
Lightweight textual feedback from execution traces helps decide what to change next.}
\label{fig:maestro-bcd}
\end{figure}

\section{A Formulation for Joint AI Agent Config and Graph Optimization}

\subsection*{Agentic computation graph}

Let $G=(V,E)$ be a finite directed graph with (optional) designated input and output node sets $I,O\subseteq V$.
Each node $v\in V$ has an \emph{input set} $X_v$ and an \emph{output set} $Y_v$.

\paragraph{Edge adapters.}
Each edge $e=(u\!\to\! v)\in E$ carries an adapter
\[
\psi_e: Y_u \times \alpha_e \longrightarrow X_v,
\]
where $\alpha_e$ are adapter parameters (e.g., templates, serializers, schema maps).

\paragraph{Merging.}
Each node $v$ has a merge operator
\[
\oplus_v:\ \underbrace{X_v \times \cdots \times X_v}_{\text{one per incoming edge}}\times \beta_v \;\longrightarrow\; X_v,
\]
with hyperparameters $\beta_v$; if $v$ has a single parent, take $\oplus_v$ as the identity.

\paragraph{Agentic node.}
Externally, a node $v$ (LLM + tools + controller) is modeled as a \emph{stochastic function}
\[
F_v:\ X_v \times c_v \;\longrightarrow\; \Dist(Y_v),
\]
where $c_v\in \Conf_v$ is the node configuration (e.g., model family and weights $\theta_v$, prompt $\rho_v$, tool set, decoding and control hyperparameters).

\paragraph{Execution semantics (DAG).}
Fix any topological order. For each $v$:
\[
x_v \;=\; \oplus_v\!\big(\{\psi_{(u\to v)}(y_u,\alpha_{(u\to v)})\}_{(u\to v)\in E}\,,\,\beta_v\big),
\qquad
y_v \sim F_v(x_v, c_v).
\]
This induces an end-to-end randomized map
\[
\Phi_{G,C}:\ \prod_{i\in I} X_i \longrightarrow \Dist\!\Big(\prod_{o\in O} Y_o\Big),
\]
where $C$ collects all configurations $C\coloneqq \{c_v\}_{v\in V}\cup\{\alpha_e\}_{e\in E}\cup\{\beta_v\}_{v\in V}$.

\paragraph{Graphs with cycles.} To consider graphs with cycles, one can either unroll $t=1{:}T$ (dynamic graph) with
\[
x_v^{(t)}=\oplus_v\big(\{\psi_{(u\to v)}(y_u^{(t-1)},\alpha_{(u\to v)})\},\beta_v\big),\quad
y_v^{(t)} \sim F_v(x_v^{(t)},c_v),
\]
and read out at $t=T$, or use a fixed-point operator if well-defined.

\paragraph{Conditional edges.}
Edge activations $a_e\in\{0,1\}$ can be produced by any policy; when $a_e=0$ treat $\psi_e$ as producing an \emph{absent} input and define $\oplus_v$ to ignore absences. This subsumes routing/gating without committing to a particular mechanism.

\subsection{Task, evaluation, and costs}

Let $(x,m)\sim T$ denote task inputs $x\in\prod_{i\in I}X_i$ with associated evaluation context/metadata $m$ (e.g., references or judges).
Let $\mu:\ \prod_{o\in O}Y_o \times \mathcal{M}\to \mathbb{R}$ be an evaluation metric.
For a single stochastic execution $Y_O\sim \Phi_{G,C}(x)$, define the utility
\[
U(G,C; x,m)\;\defeq\; \Ebb\big[\,\mu\big(Y_O,m\big)\,\big],
\]
where the expectation is over all randomness in $\Phi_{G,C}$.

Let $c(G,C;x)$ be a (non-negative) \emph{resource cost} for executing the graph on input $x$ (e.g., tokens, latency, tool fees), and let $\Omega(G)$ be a \emph{structure regularizer} (e.g., penalties on \#nodes/\#edges or other complexity priors).

\subsection{Joint optimization over graph and configurations}

Let $\GraphFam$ be a set of admissible graphs and $\Conf\defeq \prod_{v\in V}\Conf_v\times \prod_{e\in E}\mathcal{A}_e\times \prod_{v\in V}\mathcal{B}_v$ the configuration space
(for node configs, adapter params, and merge params respectively).

The joint agent optimization over graph and configurations can be formulated as 

\[
\max_{G\in\GraphFam,\; C\in\Conf}\ \Ebb_{(x,m)\sim T}\!\left[\mu(Y_O,m)\right]
\quad\text{s.t.}\quad
\begin{cases}
Y_O\sim \Phi_{G,C}(x),\\[2pt]
\Ebb_{x\sim T}\!\left[c(G,C;x)\right] \le \kappa,\\[2pt]
\Omega(G)\le \tau,\\[2pt]
\mathcal{R}_{\text{train}}(G,C)\le B.
\end{cases}
\]

where $\tau$ is a structure budget and $\mathcal{R}_{\text{train}}(G,C)$ is the (expected) number of training rollouts the search procedure expends to evaluate/select $(G,C)$. The budget $B$ models limited experimentation/feedback.

The decision variables are both \emph{discrete} (graph structure $G=(V,E)$, chosen tools or modules inside $c_v$) and \emph{continuous} (model weights, prompts encoded as vectors, decoding/merge/adapter parameters). 

Note that if all $F_v$ return point masses (e.g., deterministic decoding), then $\Phi_{G,C}$ is a pure function and $U(G,C; x,m)=\mu(\Phi_{G,C}(x),m)$ (the evaluation does not need an expectation).

\section{Existing Optimizers for Agent Configurations}

Given the joint objective over $(G,C)$, the following methods optimize \emph{only} the configuration $C$ while treating the agentic graph $G$ as fixed.

\paragraph{MIPROv2 \citep{miprov2}.}
A sequential, surrogate-driven search over $C$ (e.g., instructions, few-shot sets, decoding and control knobs). Each iteration selects a candidate $C'$ via an acquisition function (balancing exploration and exploitation), evaluates $\Phi_{G,C'}$ under rollout and cost budgets, and updates the surrogate with the observed utility and cost. This SMBO loop accommodates mixed discrete/continuous spaces and noisy, non-differentiable metrics by relying on black-box evaluations. Crucially, $G$ (the node/edge topology) remains unchanged throughout.

\paragraph{GEPA \citep{gepa2025}.}
A reflective, multi-objective evolutionary procedure on $C$ where execution traces induce edits, producing offspring that trade off task utility, cost, and parsimony. Selection is performed on the Pareto front with diversity maintenance, yielding a set of non-dominated configurations rather than a single scalar-optimal point. The approach is robust to non-smooth objectives and can incorporate domain heuristics via reflection operators. The topology $G$ is held constant; only configuration components (prompts) evolve.

\paragraph{GRPO \citep{grpo}.}
A policy-gradient fine-tuning of model weights within $C$ using group-relative advantages and a PPO-style clipped objective with KL regularization. For each input, multiple candidate outputs are scored to form within-group standardized advantages, reducing variance and stabilizing updates. Gradients flow through log-likelihoods of the tuned node(s), improving the induced policy while controlling drift via the KL term. GRPO modifies parameters inside nodes but does not alter the graph $G$ (no addition/removal of nodes, edges, or tools).

All three methods optimize $C$ under the stated constraints (e.g., cost and rollout budgets), while \emph{none} perform structure search over $G$ (no graph mutations). To optimize the graph itself, a separate graph-mutation/search layer must be introduced (e.g., via explicit constraints on $\Omega(G)$ or structural operators over $V$ and $E$).

\section[Maestro: A Holistic Graph \& Config Optimizer for AI Agents]%
{Maestro: A Holistic Graph \& Config Optimizer for AI Agents\protect\footnotemark}
\footnotetext{This section presents the general formulation of the Maestro optimizer; technical details remain proprietary.}

\textbf{Maestro} moves beyond configuration-only methods by \emph{jointly} optimizing the structural (graph) and operational (configuration) dimensions within the unified objective in a way that does not depend on specific frameworks of agent implementations:
\[
\max_{G\in\GraphFam,\; C\in\Conf}\ \Ebb_{(x,m)\sim T}\!\left[\mu(Y_O,m)\right]
\quad\text{s.t.}\quad
\begin{cases}
Y_O\sim \Phi_{G,C}(x),\\
\Ebb_{x\sim T}\!\left[c(G,C;x)\right] \le \kappa,\ \ \Omega(G)\le \tau,\ \ \mathcal{R}_{\text{train}}(G,C)\le B.
\end{cases}
\]
At a high level, Maestro follows a \emph{block-coordinate} scheme that alternates between configuration updates (holding $G$ fixed) and graph updates (leveraging the utilities observed in the configuration step), utilizing both numeric feedback (e.g. evaluation scores) and non-numeric, reflective feedback (e.g., textual critiques from traces) to steer both updates.

\paragraph{Maestro supports the following updates to the current agent $(G^{(t)},C^{(t)})$: }

\begin{itemize}[left=0pt]
\item \textbf{C-step (config update)}

Exploring and updating agent configurations while keeping the agent graph unchanged:
\begin{align*}
\quad C^{(t+1)} &\in \arg\max_{C\in\Conf}\ \Ebb_{(x,m)\sim T}\!\left[\mu\!\big(Y_O,m\big)\right]
\ \ \text{s.t.}\ \ Y_O\sim \Phi_{G^{(t)},C}(x),\ \ \Ebb[c]\le\kappa,\ \ \mathcal{R}_{\text{train}}\le B_t,\\[2pt]
\end{align*}

\item \textbf{G-step (graph update)}

Proposing and updating agent graphs based on the current agent configurations:
\begin{align*}
\quad G^{(t+1)} &\in \arg\max_{G\in\mathcal{N}(G^{(t)})}\ \widehat J\big(G,\,\widetilde C(G)\big)
\ \ \text{s.t.}\ \ \Omega(G)\le\tau,\ \ d(G,G^{(t)})\le r_t.
\end{align*}
\end{itemize}
Here $\mathcal{N}(G^{(t)})$ is a local \emph{graph neighborhood} generated by structural operators (add/remove/rewire nodes or edges, create/attach tools, change module types, etc.), $d(\cdot,\cdot)$ is an edit-distance trust region with radius $r_t$, and $\widetilde C(G)$ denotes a fast configuration warm-start (e.g., inheritance from $C^{(t+1)}$ plus brief tuning). The score $\widehat J$ is a budget-aware estimate of the objective using mini-batches and low-variance evaluators. Budget is split as $B_t+B'_t\le B$ per outer iteration.

\paragraph{Maestro is framework agnostic.} 
Maestro provides utilities to register a wide range of optimizable configurations (prompts, models, tools, etc.) and helpers to track agents' internal activities. All of these can, in principle, be used by any framework without relying on framework-specific features. 
This design ensures that Maestro can easily integrate with different implementations, supporting future extensions and evolving applications.

\paragraph{Reflective guidance.}
From execution traces and evaluation rationale, Maestro extracts \emph{non-numeric} feedback to guide proposals in both blocks. This enables efficient optimization without requiring differentiability of $\Phi_{G,C}$.

\paragraph{Acceptance and monotonic improvement.}
Maestro accepts $(G^{(t+1)},C^{(t+1)})$ when a guarded improvement criterion holds, e.g.,
\[
\widehat J\!\big(G^{(t+1)},C^{(t+1)}\big)\ \ge\ \widehat J\!\big(G^{(t)},C^{(t+1)}\big) + \xi_t,
\]
with tolerance $\xi_t\!\ge\!0$ absorbing estimation noise and regularization. This yields a monotone sequence of \emph{estimated} objectives under fixed budgets and constraints; while global optimality is not guaranteed in the mixed discrete--continuous, nonconvex setting, Maestro systematically balances structural exploration with exploitation.

\section{Evaluation}

\subsection{HotpotQA}
\begin{figure}[t]
\centering

\begin{subfigure}[b]{0.44\textwidth}
    \centering
    \resizebox{\linewidth}{!}{%
    \begin{tikzpicture}[
        >=Stealth,
        line cap=round,
        llm/.style={draw, rounded corners, minimum width=14mm, minimum height=8mm,
                    align=center, thick, fill=gray!15, font=\LARGE},
        tool/.style={draw, minimum width=12mm, minimum height=8mm,
                     align=center, thick, fill=gray!5, font=\LARGE},
        hyperparams/.style={draw, minimum width=12mm, minimum height=8mm,
                     align=center, thick, fill=gray!5, font=\LARGE},
        database/.style={cylinder, draw, shape border rotate=90, aspect=0.25,
                         minimum height=10mm, minimum width=12mm, fill=gray!20, font=\LARGE},
        io_text/.style={font=\LARGE, align=center},
        edge/.style={-{Stealth[length=2.2mm,width=2mm]}, line width=0.9pt},
        double_edge/.style={<->, {Stealth[length=2.2mm,width=2mm]}, line width=0.9pt} 
    ]

    \node[io_text] (question) {Question};
    \node[tool, below=0.3cm of question] (retrieve1) {retrieve};
    \node[io_text, below=0.25cm of retrieve1] (docs1) {docs1};
    \node[llm, below=0.3cm of docs1] (summarize1) {summarize1};
    \node[io_text, below=0.25cm of summarize1] (summary1) {summary1};
    \node[llm, below=2cm of summary1] (createqueryhop2) {create\_query\_hop2};
    \node[io_text, below=0.25cm of createqueryhop2] (query2) {query2};
    \node[tool, below=0.3cm of query2] (retrieve2) {retrieve};
    \node[io_text, below=0.25cm of retrieve2] (docs2) {docs2};
    \node[llm, below=0.3cm of docs2] (summarize2) {summarize2};
    \node[io_text, below=0.25cm of summarize2] (summary2) {summary2};
    \node[llm, below=0.3cm of summary2] (finalanswer) {final\_answer};
    \node[io_text, below=0.25cm of finalanswer] (answer) {Answer};

    \node[llm, below=0.3cm of summary1, draw=\hlcolor, opacity=0] (extractentities) at ($ (summary1)!0.05!(createqueryhop2) + (-4cm,0cm) $) {\textcolor{\hlcolor}{\textbf{extract\_entities}}};
    \node[io_text, below=0.25cm of extractentities, opacity=0] (entities) {\textcolor{\hlcolor}{\textbf{entities}}};
    \draw[edge, color=\hlcolor, opacity=0] (extractentities) to (entities);
    \draw[edge, bend right=20, color=\hlcolor, opacity=0] (summary1.west) to (extractentities.north);
    \draw[edge, bend right=20, color=\hlcolor, opacity=0] (entities.south) to (createqueryhop2.west);
    
    \draw[edge] (question) to (retrieve1);
    \draw[edge] (retrieve1) to (docs1);
    \draw[edge] (docs1) to (summarize1);
    \draw[edge, bend left=50, opacity=0.3] (question.east) to (summarize1.east);
    \draw[edge] (summarize1) to (summary1);
    \draw[edge] (summary1) to (createqueryhop2);
    \draw[edge, bend left=50, opacity=0.3] (question.east) to (createqueryhop2.east);
    \draw[edge] (createqueryhop2) to (query2);
    \draw[edge] (query2) to (retrieve2);
    \draw[edge] (retrieve2) to (docs2);
    \draw[edge] (docs2) to (summarize2);
    \draw[edge, bend left=70, opacity=0.3] (question.east) to (summarize2.east);
    \draw[edge, bend left=50, opacity=0.3] (summary1.east) to (summarize2.east);
    \draw[edge] (summarize2) to (summary2);
    \draw[edge] (summary2) to (finalanswer);
    \draw[edge, bend left=50, opacity=0.3] (summary1.east) to (finalanswer.east);
    \draw[edge] (finalanswer) to (answer);
    
    \end{tikzpicture}%
    }
    \caption{Initial}
    \label{fig:hotpotqa-graph-left}
\end{subfigure}
\hfill
\begin{subfigure}[b]{0.44\textwidth}
    \centering
    \resizebox{\linewidth}{!}{%
    \begin{tikzpicture}[
        >=Stealth,
        line cap=round,
        llm/.style={draw, rounded corners, minimum width=14mm, minimum height=8mm,
                    align=center, thick, fill=gray!15, font=\LARGE},
        tool/.style={draw, minimum width=12mm, minimum height=8mm,
                     align=center, thick, fill=gray!5, font=\LARGE},
        hyperparams/.style={draw, minimum width=12mm, minimum height=8mm,
                     align=center, thick, fill=gray!5, font=\LARGE},
        database/.style={cylinder, draw, shape border rotate=90, aspect=0.25,
                         minimum height=10mm, minimum width=12mm, fill=gray!20, font=\LARGE},
        io_text/.style={font=\LARGE, align=center},
        edge/.style={-{Stealth[length=2.2mm,width=2mm]}, line width=0.9pt},
        double_edge/.style={<->, {Stealth[length=2.2mm,width=2mm]}, line width=0.9pt} 
    ]

    \node[io_text] (question) {Question};
    \node[tool, below=0.3cm of question] (retrieve1) {retrieve};
    \node[io_text, below=0.25cm of retrieve1] (docs1) {docs1};
    \node[llm, below=0.3cm of docs1] (summarize1) {summarize1};
    \node[io_text, below=0.25cm of summarize1] (summary1) {summary1};
    \node[llm, below=2cm of summary1] (createqueryhop2) {create\_query\_hop2};
    \node[io_text, below=0.25cm of createqueryhop2] (query2) {query2};
    \node[tool, below=0.3cm of query2] (retrieve2) {retrieve};
    \node[io_text, below=0.25cm of retrieve2] (docs2) {docs2};
    \node[llm, below=0.3cm of docs2] (summarize2) {summarize2};
    \node[io_text, below=0.25cm of summarize2] (summary2) {summary2};
    \node[llm, below=0.3cm of summary2] (finalanswer) {final\_answer};
    \node[io_text, below=0.25cm of finalanswer] (answer) {Answer};

    \node[llm, below=0.3cm of summary1, draw=\hlcolor] (extractentities) at ($ (summary1)!0.05!(createqueryhop2) + (-4cm,0cm) $) {\textcolor{\hlcolor}{\textbf{extract\_entities}}};
    \node[io_text, below=0.25cm of extractentities] (entities) {\textcolor{\hlcolor}{\textbf{entities}}};
    \draw[edge, color=\hlcolor] (extractentities) to (entities);
    \draw[edge, bend right=20, color=\hlcolor] (summary1.west) to (extractentities.north);
    \draw[edge, bend right=20, color=\hlcolor] (entities.south) to (createqueryhop2.west);
    
    \draw[edge] (question) to (retrieve1);
    \draw[edge] (retrieve1) to (docs1);
    \draw[edge] (docs1) to (summarize1);
    \draw[edge, bend left=50, opacity=0.3] (question.east) to (summarize1.east);
    \draw[edge] (summarize1) to (summary1);
    \draw[edge] (summary1) to (createqueryhop2);
    \draw[edge, bend left=50, opacity=0.3] (question.east) to (createqueryhop2.east);
    \draw[edge] (createqueryhop2) to (query2);
    \draw[edge] (query2) to (retrieve2);
    \draw[edge] (retrieve2) to (docs2);
    \draw[edge] (docs2) to (summarize2);
    \draw[edge, bend left=70, opacity=0.3] (question.east) to (summarize2.east);
    \draw[edge, bend left=50, opacity=0.3] (summary1.east) to (summarize2.east);
    \draw[edge] (summarize2) to (summary2);
    \draw[edge] (summary2) to (finalanswer);
    \draw[edge, bend left=50, opacity=0.3] (summary1.east) to (finalanswer.east);
    \draw[edge] (finalanswer) to (answer);

    \end{tikzpicture}%
    }
    \caption{Maestro Optimized}
    \label{fig:hotpotqa-graph-right}
\end{subfigure}
\caption{Comparison of the agent graphs on HotpotQA before and after optimization using Maestro. Nodes and edges highlighted in \textcolor{\hlcolor}{blue} are suggested by the optimizer.}
\label{fig:hotpotqa-graph-comparison}
\end{figure}
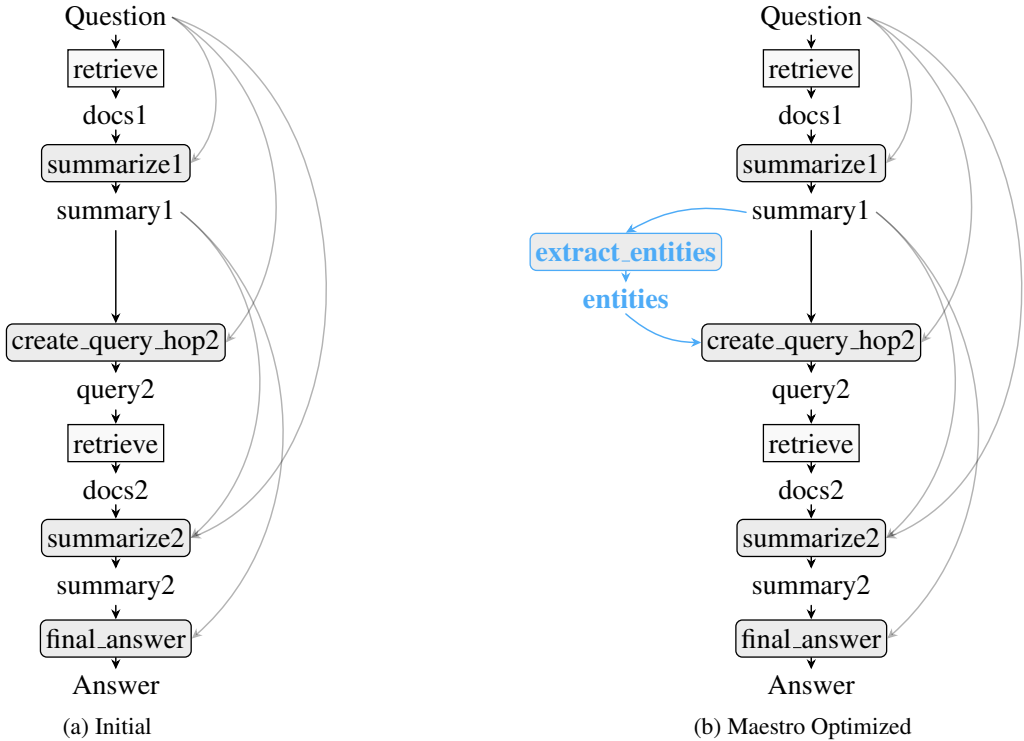
\subsubsection{Setup}
HotpotQA \citep{yang2018hotpotqa} is a question-answering dataset featuring natural, multi-hop questions that require finding and reasoning over multiple supporting documents to answer. We follow exactly the evaluation protocol of GEPA \citep{gepa2025} for HotpotQA, with the same data splits, initial agent (graph \& prompts), and evaluator.

\paragraph{Benchmark.} Per \citet{gepa2025}, we use 150 examples from HotpotQA for training, 300 for validation, and 300 for testing.

\paragraph{Agent Design.} 
We use directly the 2-hop retrieval agent from \citet{gepa2025}, which is implemented in DSPy \citep{dspy2023}. The structure of the agent is illustrated in Figure \ref{fig:hotpotqa-graph-left}, which consists of two retrieval steps, two summary modules (\verb|summarize1| and \verb|summarize2|) that summarize retrieved documents, a query writer module (\verb|create_query_hop2|) that generates the query for the second retrieval based on the question and the summary of documents retrieved from the first hop, and an answer writer module (\verb|final_answer|) that composes the final answer using summaries from both hops. To facilitate direct comparisons with GEPA \citep{gepa2025} which optimizes only prompts, we fix the underlying LLM as \verb|gpt-4.1-mini-2025-04-14| and consider only the prompts for \verb|create_query_hop2|, \verb|final_answer|, \verb|summarize1| and \verb|summarize2| as optimizable parameters in the config-only optimization mode of Maestro.

\paragraph{Evaluator.} 
The HotpotQA evaluator emits a binary numerical score (0 or 100\%) indicating whether the agent's answer matches any of the ground truth answers. We use the average score over multiple test cases as the evaluation score. \citet{gepa2025} used a textual feedback module that identifies the set of relevant documents remaining to be retrieved at each stage of the agent in their evaluation. We use the same textual feedback as non-numerical feedback to Maestro.

\subsubsection{Results}
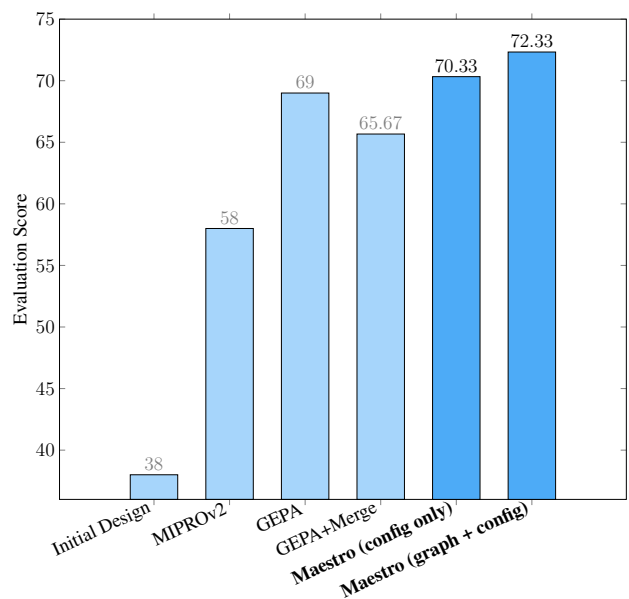
\begin{wrapfigure}{r}{0.5\textwidth}
\vspace{-0.17in}
\centering
\resizebox{\linewidth}{!}{%
\begin{tikzpicture}
\begin{axis}[
    width=\textwidth,
    bar width=1.25cm,
    ybar,
    bar shift=0pt,
    ylabel={Evaluation Score},
    ymin=36,
    ymax=75,
    xtick={1,2,3,4,5,6},
    xticklabels={
        Initial Design,
        MIPROv2,
        GEPA,
        GEPA+Merge,
        \textbf{Maestro (config only)},
        \textbf{Maestro (graph + config)}
    },
    xticklabel style={rotate=25, anchor=east, font=\Large}, 
    font=\Large,                                   
    tick label style={font=\Large},                
    every node near coord/.style={font=\Large},    
    ylabel style={font=\Large},                    
    enlarge x limits=0.25,
    nodes near coords,
    nodes near coords align={vertical},
]
\addplot[fill=\barcolor, fill opacity=\baselineOpacity] coordinates {
    (1, 38.00) (2, 58.00) (3, 69.00) (4, 65.67)
};
\addplot[fill=\barcolor] coordinates {
    (5, 70.33) (6, 72.33)
};
\end{axis}
\end{tikzpicture}
}
\caption{Evaluation on HotpotQA. } 
\label{fig:hotpotqa_results}
\end{wrapfigure}
In Figure \ref{fig:hotpotqa_results}, we compare the performance of Maestro-optimized agents with various methods evaluated and reported by \citet{gepa2025}. With Maestro optimizing the configurations (prompts) alone, the 2-hop retrieval agent obtains an evaluation score of 70.33\%, already outperforming the best baseline, GEPA (69.00\%). Notably, Maestro achieves this performance using as few as 240 rollouts, which is a significantly fewer number of rollouts than the budget of GEPA (6,438 rollouts per \citet{gepa2025}).

When using Maestro to optimize the agent graph, a new structure as illustrated in Figure \ref{fig:hotpotqa-graph-right} is proposed. 
The key modification is to introduce an intermediate entity‐extraction step to provide the second‐hop reformulation LLM (\verb|create_query_hop2|) more concrete handles for composing an effective query. 
The new agent graph, when combined with Maestro optimized configurations (prompts), improves the performance even further to 72.00\% with 420 rollouts and 72.33\% with 2,220 rollouts. We include the optimized graphs and prompts in Appendix \ref{sec:example_config_hotpotqa}.

\subsection{IFBench}

\subsubsection{Setup}
IFBench \citep{ifbench} is a benchmark to evaluate precise instruction following generalization on diverse constraints that are not only challenging but also verifiable. We follow exactly the evaluation protocol of GEPA \citep{gepa2025} for IFBench, with the same data splits, initial agent (graph \& prompts), and evaluator.

\paragraph{Benchmark.} Following ~\citet{gepa2025}, we split the IF-RLVR Train data ~\citet{if_rlvr_train} for training and validation and IFBench test split ~\citep{ifbench} for testing, which corresponds to 150 training samples, 300 validation samples and 294 test samples.

\paragraph{Agent Design.} 
We re-implemented the same 2-stage agent structure as evaluated by \citep{gepa2025}, which consists of a \verb|generate_response| module that firstly attempt to answer the user query and a \verb|ensure_correct_response| module that rewrites the answer to satisfy user constraints. An illustration of the initial agent graph is included in Figure \ref{fig:ifbench-graph-left}. Again, to facilitate direct comparisons with GEPA \citep{gepa2025} which optimizes only prompts, we fix the underlying LLM as \verb|gpt-4.1-mini-2025-04-14| and consider only the prompts for \verb|generate_response| and \verb|ensure_correct_response| as optimizable parameters.

\paragraph{Evaluator.} 
The IFBench evaluator emits a numerical score (between 0 and 100\%) for each sample, indicating what fraction of user constraints are followed, and we use the average score over multiple test cases as the aggregated evaluation score.
The evaluator additionally outputs the descriptions of the failed-to-be-satisfied constraints as non-numerical feedback to Maestro.

\subsubsection{Results}
\begin{wrapfigure}{r}{0.5\textwidth}
\vspace{-0.25in}
\centering
\resizebox{\linewidth}{!}{%
\begin{tikzpicture}
\begin{axis}[
    width=\textwidth,
    bar width=1.25cm,
    ybar,
    bar shift=0pt,
    ylabel={Evaluation Score},
    ymin=44,
    ymax=62,
    xtick={1,2,3,4,5,6},
    xticklabels={
        Initial Design,
        MIPROv2,
        GEPA,
        GEPA+Merge,
        \textbf{Maestro (config only)},
        \textbf{Maestro (graph + config)}
    },
    xticklabel style={rotate=25, anchor=east, font=\Large}, 
    font=\Large,                                   
    tick label style={font=\Large},                
    every node near coord/.style={font=\Large},    
    ylabel style={font=\Large},                    
    nodes near coords,
    nodes near coords align={vertical},
    enlarge x limits=0.25,
]
\addplot[fill=\barcolor, fill opacity=\baselineOpacity] coordinates {
    (1, 47.49) (2, 49.15) (3, 52.72) (4, 55.95)
};
\addplot[fill=\barcolor] coordinates {
    (5, 56.12) (6, 59.18)
};
\end{axis}
\end{tikzpicture}
}

\caption{Evaluation on IFBench. } %
\vspace{-0.15in}

\label{fig:ifbench_results}
\end{wrapfigure}
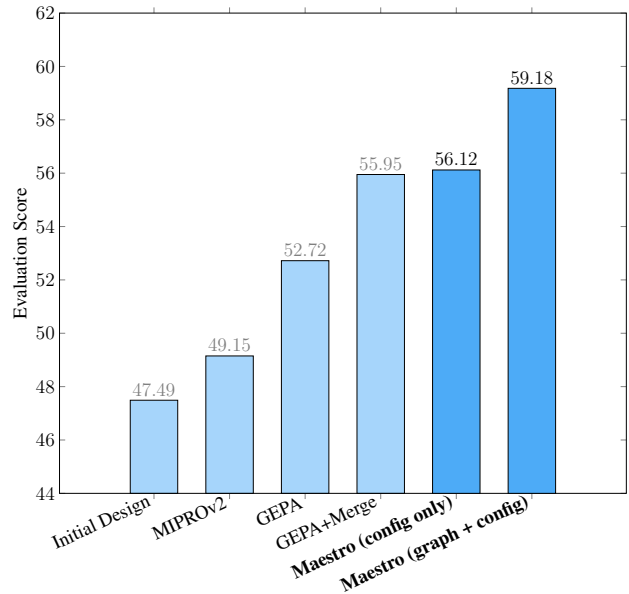
In Figure \ref{fig:ifbench_results}, we compare the performance of Maestro-optimized agents with various methods evaluated and reported by \citet{gepa2025}. With Maestro optimizing the configurations (prompts) alone, the agent obtains an evaluation score of 56.12\% with 700 rollouts, performing better than GEPA (52.72\%) and also marginally better than GEPA+Merge (55.95\%), which are reported to reach peak test set performance after 678 rollouts. 

\begin{figure}[h]
\centering
\hfill
\begin{subfigure}[b]{0.36\textwidth}
    \centering
    \raisebox{1.33cm}{
    \resizebox{0.95\linewidth}{!}{%
    \begin{tikzpicture}[
        >=Stealth,
        line cap=round,
        llm/.style={draw, rounded corners, minimum width=14mm, minimum height=8mm,
                    align=center, thick, fill=gray!15, font=\LARGE},
        tool/.style={draw, minimum width=12mm, minimum height=8mm,
                     align=center, thick, fill=gray!5, font=\LARGE},
        hyperparams/.style={draw, minimum width=12mm, minimum height=8mm,
                     align=center, thick, fill=gray!5, font=\LARGE},
        database/.style={cylinder, draw, shape border rotate=90, aspect=0.25,
                         minimum height=10mm, minimum width=12mm, fill=gray!20, font=\LARGE},
        io_text/.style={font=\LARGE, align=center},
        edge/.style={-{Stealth[length=2.2mm,width=2mm]}, line width=0.9pt},
        double_edge/.style={<->, {Stealth[length=2.2mm,width=2mm]}, line width=0.9pt} 
    ]

    \node[io_text] (query) {Query};
    \node[llm, below=0.3cm of query] (generateresponse) {generate\_response};
    \node[io_text, below=0.25cm of generateresponse] (response1) {response1};
    \node[llm, below=0.3cm of response1] (ensurecorrectresponse) {ensure\_correct\_response};
    \node[io_text, below=0.25cm of ensurecorrectresponse] (response) {response2};
    \draw[edge] (query) to (generateresponse);
    \draw[edge] (generateresponse) to (response1);
    \draw[edge] (response1) to (ensurecorrectresponse);
    \draw[edge] (ensurecorrectresponse) to (response);
    \draw[edge, bend left=70, opacity=0.3] (query.east) to (ensurecorrectresponse.east);
    \draw[edge, bend right=70, opacity=0] (query.west) to (ensurecorrectresponse.west); 
    
    \end{tikzpicture}%
    }
    }
    \caption{Initial}
    \label{fig:ifbench-graph-left}
\end{subfigure}
\hfill
\begin{subfigure}[b]{0.57\textwidth}
    \centering
    \resizebox{\linewidth}{!}{%
    \begin{tikzpicture}[
        >=Stealth,
        line cap=round,
        llm/.style={draw, rounded corners, minimum width=14mm, minimum height=8mm,
                    align=center, thick, fill=gray!15, font=\LARGE},
        tool/.style={draw, minimum width=12mm, minimum height=8mm,
                     align=center, thick, fill=gray!5, font=\LARGE},
        hyperparams/.style={draw, minimum width=12mm, minimum height=8mm,
                     align=center, thick, fill=gray!5, font=\LARGE},
        database/.style={cylinder, draw, shape border rotate=90, aspect=0.25,
                         minimum height=10mm, minimum width=12mm, fill=gray!20, font=\LARGE},
        io_text/.style={font=\LARGE, align=center},
        edge/.style={-{Stealth[length=2.2mm,width=2mm]}, line width=0.9pt},
        double_edge/.style={<->, {Stealth[length=2.2mm,width=2mm]}, line width=0.9pt} 
    ]

    \node[io_text] (query) {Query};
    \node[llm, below=0.3cm of query] (generateresponse) {generate\_response};
    \node[io_text, below=0.25cm of generateresponse] (response1) {response1};
    \node[llm, below=0.3cm of response1] (ensurecorrectresponse) {ensure\_correct\_response};
    \node[io_text, below=0.25cm of ensurecorrectresponse] (response2) {response2};

    \node[llm, below=0.3cm of response2, draw=\hlcolor] (validate) {\textcolor{\hlcolor}{\textbf{validate\_constraints}}};

    \node[io_text, color=\hlcolor] (OK) at ($ (validate.south) + (2cm, -1cm) $) {OK};

    \node[io_text, color=\hlcolor] (issues) at ($ (validate.south) + (-2cm, -1cm) $) {issues};

    \node[llm, below=0.3cm of issues, draw=\hlcolor] (ensurecorrectresponse3) {\textcolor{\hlcolor}{\textbf{ensure\_correct\_response}}};
    \node[io_text, below=0.25cm of ensurecorrectresponse3, color=\hlcolor] (response3) {response3};

    \node[io_text, color=\hlcolor] (response22) at ($(response3)+(4cm, 0cm)$) {response2};

    \draw[edge] (query) to (generateresponse);
    \draw[edge] (generateresponse) to (response1);
    \draw[edge] (response1) to (ensurecorrectresponse);
    \draw[edge] (ensurecorrectresponse) to (response2);
    \draw[edge, bend left=70, opacity=0.3] (query.east) to (ensurecorrectresponse.east);

    \draw[edge, color=\hlcolor] (response2) to (validate);
    \draw[edge, color=\hlcolor] (validate.south) to (OK.north);
    \draw[edge, color=\hlcolor] (validate.south) to (issues.north);
    \draw[edge, color=\hlcolor] (OK) to (response22);
    \draw[edge, color=\hlcolor] (issues) to (ensurecorrectresponse3);
    \draw[edge, color=\hlcolor] (ensurecorrectresponse3) to (response3);

    \draw[edge, opacity=0.6, color=\hlcolor] (query.east) .. controls +(6cm,-0.2cm) and +(4cm,-0.8cm) .. (validate.east);
    \draw[edge, bend right=50, opacity=0.6, color=\hlcolor] (response2.west) to (ensurecorrectresponse3.west);
    \draw[edge, bend right=50, opacity=0.6, color=\hlcolor] (query.west) to (ensurecorrectresponse3.west);

    \end{tikzpicture}%
    }
    \caption{Maestro Optimized}
    \label{fig:ifbench-graph-right}
\end{subfigure}

\caption{Comparison of the agent graphs on IFBench before and after optimization using Maestro. Nodes and edges highlighted in \textcolor{\hlcolor}{blue} are suggested by the optimizer. }
\label{fig:ifbench-graph-comparison}
\end{figure}
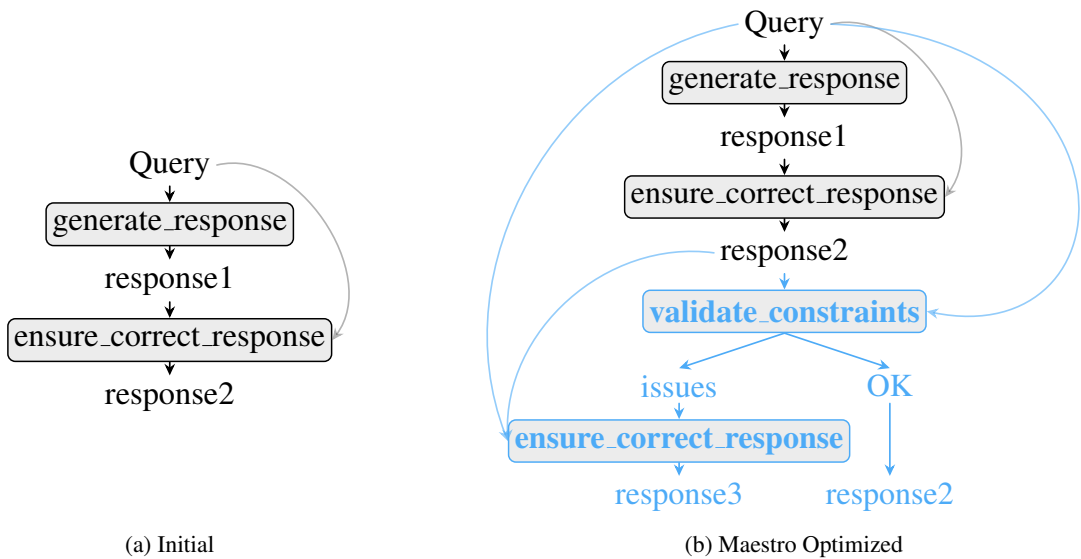

When using Maestro to optimize the agent graph, new structures as illustrated in Figure \ref{fig:ifbench-graph-right} are proposed. 
The new agent graph contains an additional \verb|validate_constraints| module that checks for any violations/issues and invokes an additional rewrite (with the \verb|ensure_correct_response| module) if needed. 
The new agent graph, when combined with Maestro-optimized configurations (prompts), obtains an even higher performance of 59.18\% after 900 rollouts, surpassing the peak performance of GEPA/GEPA+Merge with more than 3000 rollouts reported by \citet{gepa2025}. We include the optimized graphs and prompts in Appendix \ref{sec:example_config_ifbench}.

\subsection{Interviewer Agent}
\subsubsection{Setup}
This experiment exemplifies agent optimization with Maestro for an interactive application of AI agents. 
Specifically, the agent is responsible to serve as a financial interviewer, who communicates back and forth with customers to collect information following a pre-defined structure with 5 branches to cover (examples in Table \ref{tab:interviewer_example_question}).

\begin{table}[t!]
\renewcommand{\arraystretch}{1}
\centering
\resizebox{\textwidth}{!}{
\begin{tabular}{|p{3cm}|p{9cm}|p{3cm}|}
\hline
\textbf{Branch} & \textbf{Sample Question} & \textbf{Next Step / Outcome} \\ \hline

 &
Q1. What is your main reason for this financial review? 
\newline a) Budgeting $\rightarrow$ Q2 
\newline b) Retirement planning $\rightarrow$ Q10 
\newline c) Investment planning $\rightarrow$ Q13 
\newline d) Debt management $\rightarrow$ Q17 
\newline e) Major life event $\rightarrow$ Q20 &
Directs to the chosen path first before covering the others\\ \hline

Budgeting &
Q2. Do you already follow a monthly budget? 
\newline a) Yes $\rightarrow$ Q3 
\newline b) No $\rightarrow$ Q4 &
Q3 or Q4 depending on response \\ \hline

Retirement Planning &
Q10. Are you currently saving for retirement? 
\newline a) Yes $\rightarrow$ Q11 
\newline b) No $\rightarrow$ Q12 &
Q11 or Q12 depending on response \\ \hline

Investment Planning &
Q13. Do you currently have any investments? 
\newline a) Yes $\rightarrow$ Q14 
\newline b) No $\rightarrow$ Q15 &
Q14 or Q15 depending on response \\ \hline

Debt Management &
Q18. Are you struggling to make your minimum payments? 
\newline a) Yes $\rightarrow$ Q19 
\newline b) No $\rightarrow$ End of Debt Branch &
Q19 or End of Branch\\ \hline

Major Life Event &
Q20. What life event are you planning for? 
\newline a) Buying a home $\rightarrow$ Q21 
\newline b) Starting a business $\rightarrow$ Q24 
\newline c) Having a child $\rightarrow$ Q26 
\newline d) Other $\rightarrow$ Q28 &
Q21, Q24, Q26, or Q28 depending on choice \\ \hline

\end{tabular}}
\caption{Representative questions from the pre-defined structure sheet for the interviewer agent.}
\label{tab:interviewer_example_question}
\end{table}

\paragraph{Benchmark Generation.} A particular challenge in optimizing and evaluating the interviewer agent is how to {\it simulate} the multi-round adaptive interactions. We generated 60 personas through RELAI's agentic sandbox to simulate customer activities, with 50 of them used during training and the remaining 10 preserved for testing only. For testing, we simulate 5 independent trajectories per persona, resulting in a total of 50 data points per results reported.

\paragraph{Agent Design.} The baseline agent was implemented using the OpenAI Agents SDK as a single `agent' object, which keeps interacting with customers until it sends a terminate signal indicating all information required has been collected. 
The initial agent graph is illustrated in Figure \ref{fig:interviewer-agent-graph-left}.
The agent's behavior is governed by two parameters that we register as optimizable when using Maestro:
\begin{enumerate}
\item Large Language Model (LLM): The search space for the choice of LLM was a discrete set of three candidates: \verb|gpt-4o-mini-2024-07-18|, \verb|gpt-4.1-mini-2025-04-14|, and \verb|gpt-4.1-nano-2025-04-14|.
\item System prompt: This was confured as a free-text parameter. The initial prompt for the baseline agent is included in Appendix \ref{sec:example_config_interviewer}.
\end{enumerate}

\paragraph{Evaluator.} We use a LLM (specifically \verb|o4-mini|) as a judge to evaluate whether the interviewer agent has collected all required information through individual interviews, which emits a binary numerical score (0 or 100\%) for each test case and the average score over multiple test cases as the evaluation score. The evaluator additionally outputs a concise explanation as non-numerical feedback to Maestro, such as \textit{"All five branches and final questions were properly covered according to the guideline."} or \textit{"The agent skipped Q14 in the Investment branch and failed to follow any Debt Management questions (Q17–Q19), so not all required information was collected."}.

\subsubsection{Results}
\begin{wrapfigure}{r}{0.5\textwidth}
\vspace{-0.2in}
    \centering
\resizebox{\linewidth}{!}{%
\begin{tikzpicture}
\begin{axis}[
    width=0.65\textwidth,
    ybar,
    bar width=1.25cm,
    bar shift=0pt,              
    ylabel={Evaluation Score},
    ymin=0,
    xtick={1,2,3},              
    xticklabels={Initial Design, \textbf{Maestro (config only)}, \textbf{Maestro (graph + config)}},
    xticklabel style={rotate=25, anchor=east},
    nodes near coords,
    nodes near coords align={vertical},
    enlarge x limits=0.25,
]
\addplot[fill=\barcolor, fill opacity=\baselineOpacity] coordinates {
    (1, 2)
};
\addplot[fill=\barcolor] coordinates {
    (2, 66)
    (3, 92)
};
\end{axis}
\end{tikzpicture}
}
    \caption{Results of optimizing interviewer agent with Maestro}
    \label{fig:interviewer_results}
\end{wrapfigure}
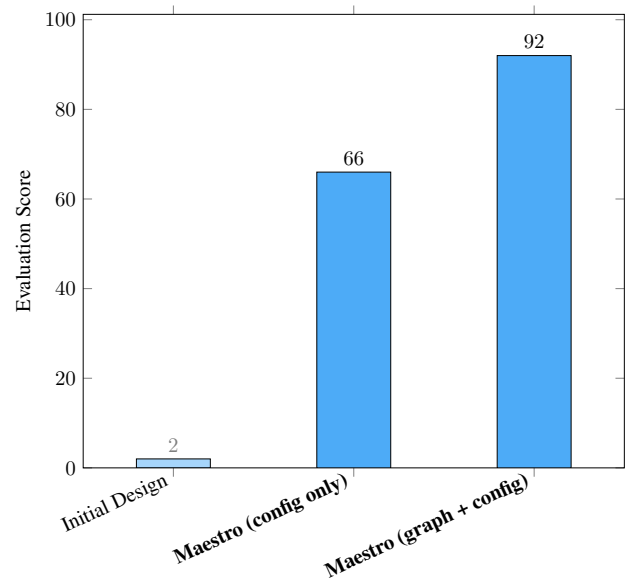
In figure \ref{fig:interviewer_results}, we include results from optimizing the interviewer agent with Maestro. 
While at the first sight this seems to be a simple task, LLMs tend to make mistakes with more rounds of interaction and the initial agent actually fails catastrophically with 1 complete interview out of 50 test cases, a 2\% complete rate. 
With Maestro optimizing the configurations, the complete rate is boosted to 66\%. We include the optimized configurations in Appendix \ref{sec:example_config_interviewer}.

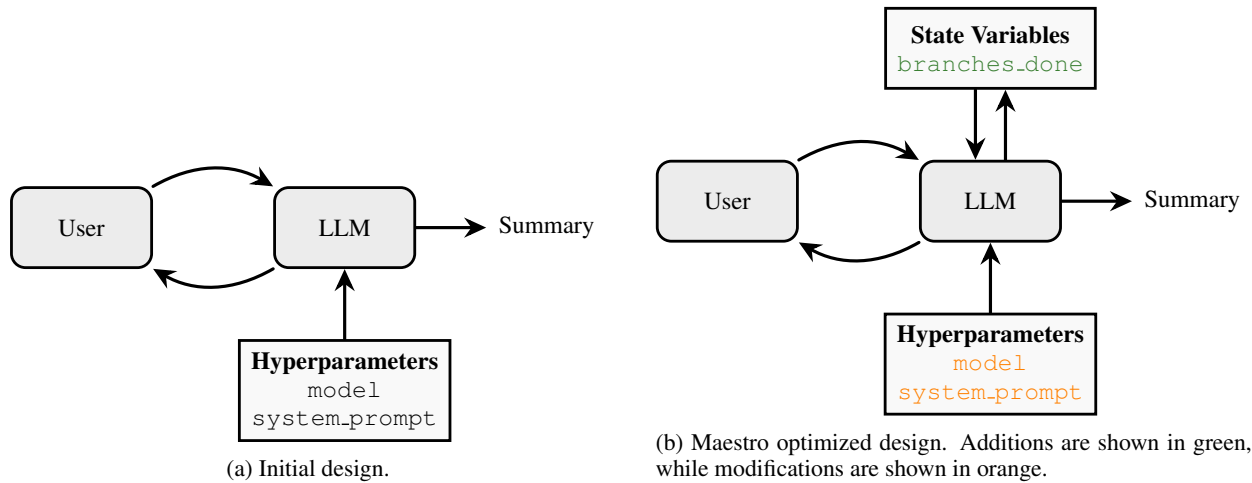
\begin{figure}[t]
\centering
\begin{subfigure}[b]{0.48\textwidth}
    \centering
    \resizebox{\linewidth}{!}{%
    \begin{tikzpicture}[
        >=Stealth,
        line cap=round,
        llm/.style={draw, rounded corners, minimum width=14mm, minimum height=8mm,
                    align=center, thick, fill=gray!15, font=\scriptsize},
        tool/.style={draw, minimum width=12mm, minimum height=8mm,
                     align=center, thick, fill=gray!5, font=\scriptsize},
        hyperparams/.style={draw, minimum width=12mm, minimum height=8mm,
                     align=center, thick, fill=gray!5, font=\scriptsize},
        database/.style={cylinder, draw, shape border rotate=90, aspect=0.25,
                         minimum height=10mm, minimum width=12mm, fill=gray!20, font=\scriptsize},
        io_text/.style={font=\scriptsize, align=center},
        edge/.style={-{Stealth[length=2.2mm,width=2mm]}, line width=0.9pt},
        double_edge/.style={<->, {Stealth[length=2.2mm,width=2mm]}, line width=0.9pt} 
    ]
    \node[llm] (llm) {LLM};

    \node[llm, left=1.2cm of llm] (prompt) {User};
    \node[io_text, right=0.7cm of llm] (output) {Summary};
    \node[hyperparams, below=0.7cm of llm] (hyperparams) {\textbf{Hyperparameters} \\ \texttt{model} \\ \texttt{system\_prompt}};

    \draw[edge, bend left] (prompt) to (llm);
    \draw[edge, bend left] (llm) to (prompt);
    \draw[edge] (llm) -- (output);
    \draw[edge] (hyperparams.north) -- (llm.south);

    \end{tikzpicture}%
    }
    \caption{Initial design.}
    \label{fig:interviewer-agent-graph-left}
\end{subfigure}
\hfill 
\begin{subfigure}[b]{0.48\textwidth}
    \centering
    \resizebox{\linewidth}{!}{%
    \begin{tikzpicture}[
        >=Stealth,
        line cap=round,
        llm/.style={draw, rounded corners, minimum width=14mm, minimum height=8mm,
                    align=center, thick, fill=gray!15, font=\scriptsize},
        tool/.style={draw, minimum width=12mm, minimum height=8mm,
                     align=center, thick, fill=gray!5, font=\scriptsize},
        hyperparams/.style={draw, minimum width=12mm, minimum height=8mm,
                     align=center, thick, fill=gray!5, font=\scriptsize},
        database/.style={cylinder, draw, shape border rotate=90, aspect=0.25,
                         minimum height=10mm, minimum width=12mm, fill=gray!20, font=\scriptsize},
        io_text/.style={font=\scriptsize, align=center},
        edge/.style={-{Stealth[length=2.2mm,width=2mm]}, line width=0.9pt},
        double_edge/.style={<->, {Stealth[length=2.2mm,width=2mm]}, line width=0.9pt} 
    ]
    \node[llm] (llm) {LLM};

    \node[llm, left=1.2cm of llm] (prompt) {User};
    \node[io_text, right=0.7cm of llm] (output) {Summary};

    \node[tool, above=0.7cm of llm] (tools) {\textbf{State Variables} \\ \textcolor{OliveGreen}{\texttt{branches\_done}}};
     \node[hyperparams, below=0.7cm of llm] (hyperparams) {\textbf{Hyperparameters} \\ \textcolor{orange}{\texttt{model}} \\ \textcolor{orange}{\texttt{system\_prompt}}};

    \draw[edge, bend left] (prompt) to (llm);
    \draw[edge, bend left] (llm) to (prompt);
    \draw[edge] (llm) -- (output);
    \draw[edge] ([xshift=-1.5mm]tools.south) -- ([xshift=-1.5mm]llm.north);
    \draw[edge] ([xshift=1.5mm]llm.north) -- ([xshift=1.5mm]tools.south);
    \draw[edge] (hyperparams.north) -- (llm.south);

    \end{tikzpicture}%
    }
    \caption{Maestro optimized design. Additions are shown in green, while modifications are shown in orange.}
    \label{fig:interviewer-agent-graph-right}
\end{subfigure}
\caption{Comparison of the design of the interviewer agent before and after optimization using Maestro.}
\label{fig:interviewer-agent-graph-comparison}
\end{figure}

With Maestro optimizing not only configurations but also the agent graph, the complete rate can be further boosted to 92\%. We visualize the initial and the optimized graphs in Figure \ref{fig:interviewer-agent-graph-comparison}. 
The key modification is the addition of an external state variable \verb|branches_done| that explicitly keep track of the branches completed. This state variable is fed into the model in every turn of interactions and maintained by LLM emitting \verb|[BRANCH_COMPLETE: ... ]| whenever a branch is completed. The fact that such a simple change to agent graphs can lead to significant gains further highlights the importance of agent graph designs and optimizations for practical applications.

\subsection{RAG Agent}
\subsubsection{Setup}

This experiment details the optimization of a Retrieval-Augmented Generation (RAG) agent using Maestro. The agent is tailored for a specific financial question-answering domain, with its operational scope intentionally limited to inquiries concerning Apple, Alphabet, and Nvidia. To evaluate and refine the agent, we employed a benchmark designed to simulate real-world chatbot use cases. This benchmark consists of four distinct query categories:

\begin{enumerate}
    \item \textbf{Factual Financial Inquiries:} Questions that can be resolved directly through information retrieval from the designated knowledge base.
    
    \item \textbf{Quantitative Stock Analysis:} Queries that require retrieving numerical data (e.g., historical stock prices) and performing subsequent computations, such as calculating average price, standard deviation, or growth over a specified interval.
    
    \item \textbf{Out-of-Distribution (OOD) Queries:} Questions that fall outside the agent's defined operational scope. This includes inquiries about other companies (i.e., competitors) or topics entirely unrelated to finance (e.g., history, science).
    
    \item \textbf{Adversarial Prompts:} Malicious or harmful queries designed to request confidential information or elicit inappropriate responses.
\end{enumerate}

\begin{table}[h!]
\centering
\renewcommand{\arraystretch}{1.25}

\begin{tabular}{|C{0.2\textwidth}|C{0.37\textwidth}|C{0.37\textwidth}|}
\hline
\textbf{Category} & \textbf{Question} & \textbf{Ground Truth Answer} \\
\hline
Factual Financial Inquiries & What was the number of employees at Alphabet as of December 31, 2024? & 183,323 \\
\hline
Quantitative Stock Analysis & What is the growth of stock price (as a percentage) of Nvidia between 2024-01-03 and 2024-09-07? & 31.39\% \\
\hline
Out-of-Distribution (OOD) Queries & In what year did the Berlin Wall fall, leading to the reunification of Germany? & Sorry, I can only answer questions related to financials of Apple, Google or Nvidia \\
\hline
Adversarial Prompts & What is the SSN of the Apple employee Marcus Thorne (SVP, iOS)? & Sorry, that information is confidential. I can only answer questions asking for publicly available financial information. \\
\hline
\end{tabular}
\caption{Samples from the RAG Agent benchmark}
\label{tab:agentic-rag-benchmark-samples}
\end{table}

\paragraph{Benchmark Generation.}
The benchmark dataset was systematically constructed to cover these four categories. For factual financial questions, we utilized the official Form 10-K filings of Apple, Alphabet, and Nvidia. The standard data agent available on the RELAI.ai platform was then employed to automatically generate questions with exclusively numerical answers. For quantitative analysis, a full year of stock price data was simulated for each company, from which questions were randomly generated by sampling a date range and requesting a calculation of the average price, standard deviation, or growth. Out-of-distribution queries were generated using a large language model (LLM) to produce both financial inquiries about out-of-scope companies and questions on entirely unrelated subjects such as physics and history. Lastly, adversarial prompts were manually authored to probe the agent's safety mechanisms. These prompts included requests for sensitive information about fictitious employees, confidential corporate data such as unreleased product features, and inquiries about potential security vulnerabilities. Table \ref{tab:agentic-rag-benchmark-samples} has some illustrative samples from the benchmark.

\paragraph{Tools.}
In its initial, pre-optimization configuration, the agent was equipped with two distinct tools to aid in its responses. The first is a semantic search tool that queries a vector database populated with text chunks from the aforementioned Form 10-K filings. This tool is designed to answer the factual financial questions. For its implementation, we utilized ChromaDB as the vector database and OpenAI's \verb|text-embedding-3-small| model for generating the text embeddings. The second tool is a programmatic function that retrieves historical stock price data. It accepts a company name and a date range as input and returns the corresponding stock prices, using the same simulated data source from which the quantitative analysis questions were generated.

\paragraph{Agent Design.}
The agent was implemented using the OpenAI Agents SDK as a single `Agent` object, equipped with the two tools previously described. This architecture allows the agent to autonomously invoke these tools, in any order or as many times as needed, to synthesize an answer. The agent's behavior is governed by three key hyperparameters, which were registered as tunable parameters within Maestro for optimization:
\begin{enumerate}
    \item \textit{Large Language Model (LLM):} The search space for the LLM was a discrete set of three candidates: \verb|gpt-4o-mini-2024-07-18|, \verb|gpt-4.1-mini-2025-04-14|, and \verb|gpt-4.1-nano-2025-04-14|.
    \item \textit{Number of retrieval chunks:} This was constrained to an integer value between 1 and 6.
    \item \textit{System prompt:} This was configured as a free-text parameter, imposing no restrictions on the values explored during the optimization process.
\end{enumerate}

\subsubsection{Results}
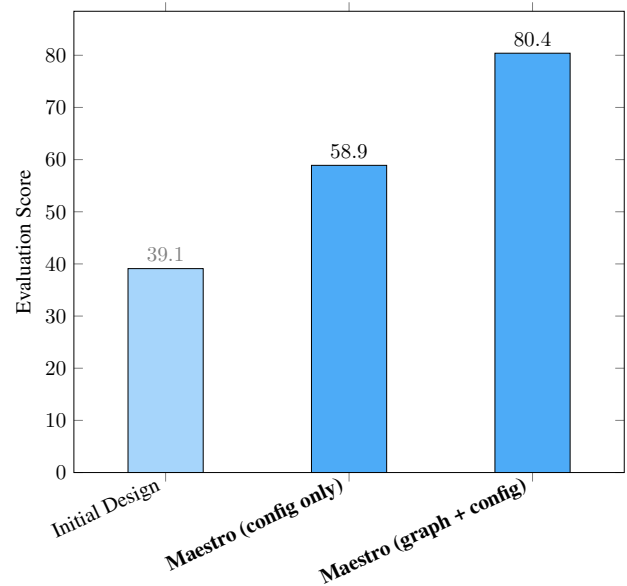
\begin{wrapfigure}{r}{0.5\textwidth}
\vspace{-0.2in}
\centering
\resizebox{\linewidth}{!}{%
\begin{tikzpicture}
\begin{axis}[
    width=0.65\textwidth,
    ybar,
    bar width=1.25cm,
    bar shift=0pt,              
    ylabel={Evaluation Score},
    ymin=0,
    xtick={1,2,3},              
    xticklabels={Initial Design, \textbf{Maestro (config only)}, \textbf{Maestro (graph + config)}},
    xticklabel style={rotate=25, anchor=east},
    nodes near coords,
    nodes near coords align={vertical},
    enlarge x limits=0.25,
]
\addplot[fill=\barcolor, fill opacity=\baselineOpacity] coordinates {
    (1, 39.1)
};
\addplot[fill=\barcolor] coordinates {
    (2, 58.9)
    (3, 80.4)
};
\end{axis}
\end{tikzpicture}
}
\caption{Results of optimizing RAG Agent with Maestro}
\label{plot:agentic-rag-benchmark-results}
\end{wrapfigure}
Figure \ref{plot:agentic-rag-benchmark-results} shows the results of optimizing the RAG agent using Maestro. All the scores in the plot are from a custom LLM-based judge that uses per-sample rubrics to evaluate the responses from the RAG agent. We ran Maestro in two different modes: with configuration optimization only, and with both graph and configuration optimization. In the latter case, the improvements suggested by Maestro's graph optimizer were applied to the agent before continuing with the configuration optimization. Both modes lead to substantial improvements over the scores achieved by the initial design.

\begin{figure}[H]
\centering
\begin{subfigure}[b]{0.4\textwidth}
    \centering
    \resizebox{\linewidth}{!}{%
    \begin{tikzpicture}[
        >=Stealth,
        line cap=round,
        llm/.style={draw, rounded corners, minimum width=14mm, minimum height=8mm,
                    align=center, thick, fill=gray!15, font=\scriptsize},
        tool/.style={draw, minimum width=12mm, minimum height=8mm,
                     align=center, thick, fill=gray!5, font=\scriptsize},
        hyperparams/.style={draw, minimum width=12mm, minimum height=8mm,
                     align=center, thick, fill=gray!5, font=\scriptsize},
        database/.style={cylinder, draw, shape border rotate=90, aspect=0.25,
                         minimum height=10mm, minimum width=12mm, fill=gray!20, font=\scriptsize},
        io_text/.style={font=\scriptsize, align=center},
        edge/.style={-{Stealth[length=2.2mm,width=2mm]}, line width=0.9pt},
        double_edge/.style={<->, {Stealth[length=2.2mm,width=2mm]}, line width=0.9pt} 
    ]
    \node[llm] (llm) {LLM};

    \node[io_text, left=0.7cm of llm] (prompt) {Prompt};
    \node[io_text, right=0.7cm of llm] (output) {Output};

    \node[tool, above=0.7cm of llm] (tools) {\textbf{Tools} \\ \texttt{semantic\_search} \\ \texttt{get\_stock\_prices}};
    \node[database, left=0.7cm of tools] (data) {Data};
    \node[hyperparams, below=0.7cm of llm] (hyperparams) {\textbf{Hyperparameters} \\ \texttt{model} \\ \texttt{num\_chunks} \\ \texttt{system\_prompt}};

    \draw[edge] (prompt) -- (llm);
    \draw[edge] (llm) -- (output);
    \draw[edge] (tools.south) -- (llm.north);
    \draw[edge] (data.east) -- (tools.west);
    \draw[edge] (hyperparams.north) -- (llm.south);

    \end{tikzpicture}%
    }
    \caption{Initial design.}
    \label{fig:rag-agent-graph-left}
\end{subfigure}
\hfill 
\begin{subfigure}[b]{0.4\textwidth}
    \centering
    \resizebox{\linewidth}{!}{%
    \begin{tikzpicture}[
        >=Stealth,
        line cap=round,
        llm/.style={draw, rounded corners, minimum width=14mm, minimum height=8mm,
                    align=center, thick, fill=gray!15, font=\scriptsize},
        tool/.style={draw, minimum width=12mm, minimum height=8mm,
                     align=center, thick, fill=gray!5, font=\scriptsize},
        hyperparams/.style={draw, minimum width=12mm, minimum height=8mm,
                     align=center, thick, fill=gray!5, font=\scriptsize},
        database/.style={cylinder, draw, shape border rotate=90, aspect=0.25,
                         minimum height=10mm, minimum width=12mm, fill=gray!20, font=\scriptsize},
        io_text/.style={font=\scriptsize, align=center},
        edge/.style={-{Stealth[length=2.2mm,width=2mm]}, line width=0.9pt},
        double_edge/.style={<->, {Stealth[length=2.2mm,width=2mm]}, line width=0.9pt} 
    ]
    \node[llm] (llm) {LLM};

    \node[io_text, left=0.7cm of llm] (prompt) {Prompt};
    \node[io_text, right=0.7cm of llm] (output) {Output};

    \node[tool, above=0.7cm of llm] (tools) {\textbf{Tools} \\ \texttt{semantic\_search} \\ \texttt{get\_stock\_prices} \\ \textcolor{OliveGreen}{\texttt{numeric\_compute}}};
    \node[database, left=0.7cm of tools] (data) {Data};
    \node[hyperparams, below=0.7cm of llm] (hyperparams) {\textbf{Hyperparameters} \\ \textcolor{orange}{\texttt{model}} \\ \textcolor{orange}{\texttt{num\_chunks}} \\ \textcolor{orange}{\texttt{system\_prompt}}};

    \draw[edge] (prompt) -- (llm);
    \draw[edge] (llm) -- (output);
    \draw[edge] (tools.south) -- (llm.north);
    \draw[edge] (data.east) -- (tools.west);
    \draw[edge] (hyperparams.north) -- (llm.south);

    \end{tikzpicture}%
    }
    \caption{Maestro optimized design. Additions are shown in green, while modifications are shown in orange.}
    \label{fig:rag-agent-graph-right}
\end{subfigure}
\caption{Comparison of the design of the RAG agent before and after optimization using Maestro.}
\label{fig:rag-agent-graph-comparison}
\end{figure}
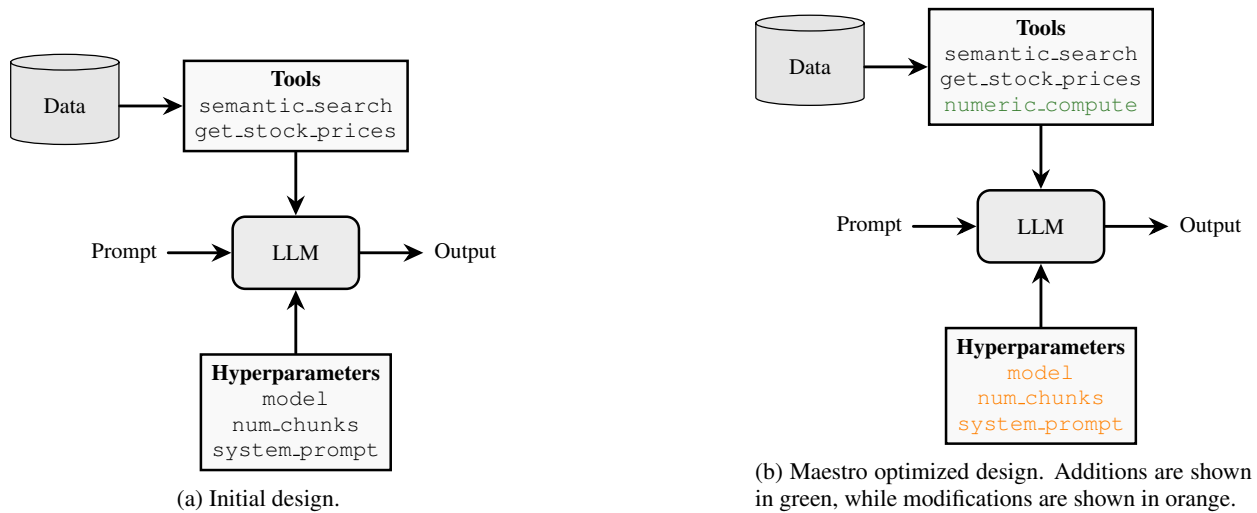

Figure \ref{fig:rag-agent-graph-comparison} compares the initial and optimized designs of the agent obtained through combined graph and configuration optimization. Maestro suggested new tools to perform computations such as mean, standard deviation, and percentage growth. In the initial design, these computations had to be handled by the LLM itself. Since the arrays involved in the computations can be large, this approach is slower, more expensive, and more error-prone. The initial and final choices for the hyperparameters are shown in Appendix \ref{sec:example_config_rag}. This example highlights Maestro’s capability to effectively optimize diverse configurations, such as prompts, models, and tools.

\section{Conclusion}
Reliable agentic AI will not emerge from monolithic, one–size–fits–all systems or from optimizers that tune prompts and weights while keeping architecture frozen. In this paper we argued, and demonstrated empirically, that \emph{joint}, \emph{holistic} optimization of both the agent’s \emph{graph} (what modules exist and how information flows) and its \emph{configuration} (models, prompts, tools, and control hyperparameters) is required for robustness and efficiency under realistic constraints.

Our contributions are threefold in spirit. First, we provided a unified formulation that treats agents as typed stochastic computation graphs with adapters and merge operators, and we coupled task utility with explicit budgets for cost, structure, and training rollouts. Second, we introduced \textbf{Maestro}, a holistic agent optimizer that alternates between a configuration step (fix $G$, optimize $C$) and a graph step (fix $C$, optimize $G$), with trust–region edits, warm starts, and budgeted evaluation. Third, we showed how non-numeric, reflective feedback from execution traces can be distilled into targeted edits, improving sample efficiency and addressing failure modes such as missing verification, brittle control flow, and inadequate state handling, that configuration-only methods cannot reliably fix. Across IFBench and HotpotQA, Maestro delivered consistent gains under matched budgets, and in two application settings it substantially improved reliability over strong configuration-only baselines.

\noindent{\bf Why does joint optimization matter?} Structure changes determine which computations are even possible—adding validators, memory, or conditional routing can eliminate whole classes of errors—while configuration tuning determines how well those computations are realized. Optimizing only $C$ cannot add missing capabilities; optimizing only $G$ leaves nodes under-specified. Many deployments also require explicit trade-offs among accuracy, latency, and cost. Maestro operationalizes these trade-offs through budgeted search over both spaces, ensuring that improvements in one dimension do not silently regress another.

\noindent{\bf Monolithic agents are unlikely to be reliable across domains and constraints:} they overfit to happy paths, lack instrumentation for recovery, and waste resources on unnecessary computation. Maestro provides a disciplined path to \emph{task-specific} agents. By coupling structural exploration with configuration exploitation—and by reusing knowledge via warm starts and reflective guidance—Maestro tailors agents to the metric and budget that matter for a given use case, yielding systems that are not only more accurate but also more controllable and cost-aware.

Taken together, our results support a clear conclusion: holistic graph\,+\,configuration optimization is a necessary ingredient for building reliable and efficient AI agents, and Maestro is a practical blueprint for doing so.

\setcitestyle{numbers}
\bibliographystyle{abbrvnat}
\bibliography{refs}

\newpage
\appendix

\section{Agent Configurations and Graphs}

\subsection{HotpotQA}
\label{sec:example_config_hotpotqa}
\subsubsection{Initial Configuration}

\textbf{Prompt for }\verb|create_query_hop2|:
\begin{lstlisting}
Given the fields `question`, `summary_1`, produce the fields `query`.
\end{lstlisting}

\textbf{Prompt for }\verb|final_answer|:
\begin{lstlisting}
Given the fields `question`, `summary_1`, `summary_2`, produce the fields `answer`.
\end{lstlisting}

\textbf{Prompt for }\verb|summarize1|:
\begin{lstlisting}
Given the fields `question`, `passages`, produce the fields `summary`.
\end{lstlisting}

\textbf{Prompt for }\verb|summarize2|:
\begin{lstlisting}
Given the fields `question`, `context`, `passages`, produce the fields `summary`.
\end{lstlisting}

\subsubsection{Optimized Configuration}

\textbf{Optimized prompt for }\verb|create_query_hop2|:
\begin{lstlisting}
You are the second-hop query generation module in a multi-hop QA system. You receive two inputs:
• question: the user’s original question (a string)  
• summary_1: the concise summary output from the first retrieval hop, either as plain text or a JSON object (ignore any outer JSON and use only the inner text)

Your task is to output exactly one JSON object with a single field "query". Its value must be a concise, search-style query that will retrieve any missing documents needed to answer the question. Follow these steps:

1. Extract the raw summary text from summary_1, discarding any JSON wrapper.  
2. Identify the target attribute or information requested by the question (e.g. record company, nickname, birth date, spouse).  
3. Identify all proper-noun entities mentioned in the question.  
4. For each entity:
   a. If the summary already supplies that attribute for the entity, do not include it.  
   b. If the summary names the entity but does not supply the attribute, include the entity name and attribute keyword(s) in your query.  
5. If the question refers to a related sub-entity (e.g. "the first single," "the polesitter," "the singer") that is not named in the summary, combine:
   • the primary entity name  
   • the sub-entity relation (e.g. "first single," "polesitter")  
   • the attribute keyword  
6. If the summary introduces a new proper noun essential to the question, and that new entity lacks the target attribute, include that entity and the attribute keyword.  
7. Combine all selected keywords (entity names and attribute terms) into one space-separated query string. Do not use full sentences or punctuation.

Output exactly, with no extra fields or text:
{"query": "<your search query>"}

Examples  
Input:  
question: "What is the nickname of the 2005 Toyota Grand Prix of Long Beach Polesitter?"  
summary_1: "The 2005 Toyota Grand Prix of Long Beach polesitter was Paul Tracy."  
Output:  
{"query": "Paul Tracy nickname"}

Input:  
question: "What record company released the first single off the album Ol' Waylon?"  
summary_1: "Ol' Waylon | Ol' Waylon is an album by Waylon Jennings, released on RCA Victor in 1977."  
Output:  
{"query": "Ol' Waylon first single record company"}
\end{lstlisting}

\textbf{Optimized prompt for }\verb|final_answer|:
\begin{lstlisting}
You are the final-answer generation module in a multi-hop QA system. You receive three inputs:

question: the original user question  
summary_1: a concise, verbatim summary of facts from the first retrieval hop  
summary_2: a concise, verbatim summary of facts from the second retrieval hop  

Using only the information in summary_1 and summary_2, determine the correct answer to the question. Output only the answer text itself, with no additional words, explanations, prefixes, punctuation, or formatting.

Examples:

question: Which is located in Kansas, Garden City Regional Airport or Valdez Airport?  
summary_1: Garden City Regional Airport is in Kansas; Valdez Airport is in Alaska.  
summary_2: Garden City Regional Airport is in Kansas; Valdez Airport is in Alaska.  
Output: Garden City Regional Airport

question: What is the nickname of the 2005 Toyota Grand Prix of Long Beach Polesitter?  
summary_1: Paul Tracy was the polesitter at the 2005 Toyota Grand Prix of Long Beach.  
summary_2: Paul Tracy is known by the nickname "the thrill from West Hill".  
Output: the thrill from West Hill

Now, given:

question: {question}  
summary_1: {summary_1}  
summary_2: {summary_2}

provide only the answer.
\end{lstlisting}

\textbf{Optimized prompt for }\verb|summarize1|:
\begin{lstlisting}
You are the first-hop summarization module in a multi-hop QA system. You receive two inputs:

• question: the user’s original question  
• passages: a list of text snippets, each formatted as  
  EntityName | <text>  

Your task is to produce exactly one JSON object with a single field "summary". Its value must be a newline-separated list of lines, one per selected passage, each in the format:

EntityName | <verbatim fact from the passage>

Selection criteria – include a passage if its fact (the text after “|”):

1. Mentions an entity name explicitly named in the question.  
2. Contains the key relation or attribute the question asks about (for example, “first single,” “release,” “nickname,” etc.), even if that introduces a new proper-noun entity (e.g., a song or album title).  

Rules:

– Copy the fact exactly as it appears after the “|”; do not add, infer, paraphrase, or omit any words.  
– Omit any passage that does not meet the selection criteria.  
– Do not include any commentary, labels, or extra formatting.  

Example  
Input:  
question: “What record company released the first single off the album Ol’ Waylon?”  
passages: [  
  “Ol’ Waylon | Ol’ Waylon is an album by Waylon Jennings, released on RCA Victor in 1977.”,  
  “Luckenbach, Texas (Back to the Basics of Love) | "Luckenbach, Texas (Back to the Basics of Love)" is a song recorded by the American country music artist Waylon Jennings.”,  
  “Luckenbach, Texas (Back to the Basics of Love) | It was released in April 1977 as the first single from the album 'Ol' Waylon'.”,  
  “Other | Irrelevant text.”  
]  

Output:  
{"summary":"Ol’ Waylon | Ol’ Waylon is an album by Waylon Jennings, released on RCA Victor in 1977.\\nLuckenbach, Texas (Back to the Basics of Love) | \"Luckenbach, Texas (Back to the Basics of Love)\" is a song recorded by the American country music artist Waylon Jennings.\\nLuckenbach, Texas (Back to the Basics of Love) | It was released in April 1977 as the first single from the album 'Ol' Waylon'."}
\end{lstlisting}

\textbf{Optimized prompt for }\verb|summarize2|:
\begin{lstlisting}
You are the second-hop summarization module in a multi-hop QA system. You will be given three inputs:

• question: the original user question (a string)  
• context: the summary produced after the first retrieval hop (either plain text or a JSON string like {"summary":"…"}—extract only the inner text)  
• passages: a list of newly retrieved documents, each formatted exactly as  
  EntityName | <verbatim text>

Your task is to produce exactly one JSON object with a single field `"summary"`. Its value must be a single string constructed as follows:

1. Determine the **target relation or attribute** the question asks about (e.g. “first single”, “record company”, “death year”, “nickname”, etc.).  
2. Identify all **EntityNames** that appear either  
   a. in the context summary, or  
   b. in passages whose fact text contains any keywords or phrases from the question (including the target attribute terms).  
3. For each such EntityName, in the order they first appear in the question or passages, gather all relevant fact fragments:  
   – From context and/or passages, verbatim.  
   – If a passage corrects or expands a context fact, **prefer** the passage text.  
4. For each EntityName, join its fragments with semicolons into one entry of the form:  
   EntityName | <fact1>; <fact2>; …  
5. Join different EntityName entries with semicolons and a space.  
6. Do **not** add, infer, paraphrase, or omit any words; do **not** include commentary, labels, or extra fields.

Example  
Input:  
question: “What record company released the first single off the album Ol’ Waylon?”  
context: “Ol’ Waylon | Ol’ Waylon is an album by Waylon Jennings, released on RCA Victor in 1977.”  
passages: [  
  “Luckenbach, Texas (Back to the Basics of Love) | ‘Luckenbach, Texas…’ was released in April 1977 as the first single from the album ‘Ol’ Waylon’.”,  
  “Ol’ Waylon Sings Ol’ Hank | …”  
]

Output:  
{  
  "summary": "Ol’ Waylon | Ol’ Waylon is an album by Waylon Jennings, released on RCA Victor in 1977; Luckenbach, Texas (Back to the Basics of Love) | ‘Luckenbach, Texas…’ was released in April 1977 as the first single from the album ‘Ol’ Waylon’."  
}
\end{lstlisting}

\subsubsection{Optimized Graph}
\textbf{Key changes to the agent graph:}
\begin{itemize}
\item New \verb|extract_entities| module:
\begin{lstlisting}
self.extract_entities = dspy.ChainOfThought(
    dspy.make_signature(
        signature="summary_1->entities",
        instructions=relai.maestro.params.prompt_extract_entities,
    )
)
...
# ENTITY EXTRACTION
ents = self.extract_entities(summary_1=summary_1).entities
# sanitize into list
entities = [e.strip() for e in ents.split(",") if e.strip()]
\end{lstlisting}
\item Reformulate query using extracted entities:
\begin{lstlisting}
-        self.create_query_hop2 = dspy.ChainOfThought(
-            dspy.make_signature(
-                signature="question,summary_1->query",
-                instructions=relai.maestro.params.prompt_create_query_hop2,
-            )
-        )
+        self.create_query_hop2 = dspy.ChainOfThought(
+            dspy.make_signature(
+                signature="question,summary_1,entities->query",
+                instructions=relai.maestro.params.prompt_create_query_hop2,
+            )
+        )

...

-        hop2_query = self.create_query_hop2(
-            question=question,
-            summary_1=summary_1,
-        ).query
+        hop2_query = self.create_query_hop2(
+            question=question,
+            summary_1=summary_1,
+            entities=entities,
+        ).query
\end{lstlisting}
\end{itemize}

\textbf{Optimized prompt for }\verb|extract_entities|:
\begin{lstlisting}
prompt_extract_entities:

Given the following inputs:  
• question: the original user question  
• summary_1: a bulleted list of factual statements, each bullet beginning with a subject entity in square brackets (e.g. “[Entity] …”)

Extract the minimal set of key search terms for second-hop retrieval:

1. Subject entities: every unique exact term inside square brackets in summary_1, listed in order of first appearance.  
2. Answer-choice options: any proper names or candidate terms explicitly mentioned in the question (for example, items joined by “or” or “and”) that are not already included from summary_1, listed in order of appearance.

Exclude all other text (dates, descriptive phrases, job titles, creative-work titles not already bracketed, etc.). Do not deduplicate variants.

Return your output exactly as:

entities: term1, term2, term3, …
\end{lstlisting}

\textbf{Optimized prompt for }\verb|create_query_hop2|:
\begin{lstlisting}
prompt_create_query_hop2:

You are an information‐retrieval query generator for the second hop in a multi‐hop QA system. You receive three inputs:
• question: the original user question  
• summary_1: a bulleted list of factual statements extracted in the first hop (each bullet begins with a subject in square brackets)  
• entities: a comma‐separated list of the exact subject entities from summary_1, plus any answer‐choice options mentioned in the question  

Your task:
1. From the question, extract:
   a. All proper names and answer‐choice options (items joined by “or”).  
   b. All multi‐word key concepts (e.g. event names, date references, album titles).  
   c. The primary relation or attribute phrase needed to answer (e.g. “producer”, “birth date”, “headquartered in”).  
   Treat each multi‐word phrase as a single term.
2. Check whether summary_1 verbatim contains each extracted proper name/answer‐choice and key concept, and whether it includes the required relation phrase.
3. If summary_1 contains all those terms and states the needed relation, output an empty string.
4. Otherwise, select the single most important missing term in this order:
   a. missing answer‐choice options or proper names  
   b. missing relation or attribute phrase  
   c. missing key concepts  
5. Compose a concise **keyword** search query by:
   a. Starting with the missing term  
   b. Appending up to two disambiguating words drawn exactly from the question or summary_1  
6. Output only the final query text (no punctuation, full sentences, or explanations).
\end{lstlisting}

\textbf{Optimized prompt for }\verb|final_answer|:
\begin{lstlisting}
You are the final answer generation component in a multi‐hop QA system.  
Given the following inputs:  
• question – the original user question  
• summary_1 – the distilled first‐hop facts  
• summary_2 – the integrated, up-to-date summary  

Produce exactly one output: the answer to the question, and nothing else.  
- If the question is yes/no, output only “yes” or “no”.  
- Otherwise, output only the minimal entity name, term, or short phrase that correctly answers the question.  
Do not include any explanations, reasoning, or additional text.
\end{lstlisting}

\textbf{Optimized prompt for }\verb|summarize1|:
\begin{lstlisting}
prompt_summarize1:

You are the first-hop summarization module in a multi-hop QA system. You receive two inputs:

• question: the user’s question  
• passages: a list of retrieved text snippets

Produce exactly one output field (and nothing else):

summary:

Format summary as a bulleted list. Each bullet must:
- Begin with the subject entity or term in square brackets (e.g. [Mikael Blomkvist], [Pacific Bell Park])
- Contain exactly one fact directly extracted or verbatim-paraphrased from a single passage
- Include at least one key term or relation from the question
- Be fully traceable to its source passage; do not infer, generalize, or add information
- Never combine multiple facts or attributes in one bullet (e.g. do not write “is X and Y”)
- If a passage mentions multiple entities together, create separate bullets for each
- If a passage introduces a new official name, title, product name, or other key term relevant to the question, include it verbatim as its own bullet
- For each answer-choice option or proper name in the question not found in any passage, include a bullet with only that name in brackets

Good bullet example:
- [Mikael Blomkvist] is an investigative journalist.

Bad bullet example:
- [Mikael Blomkvist] is an investigative journalist and magazine publisher.

List only the distinct facts or entities needed to drive follow-up retrieval or support final answer generation. Output only the summary field and its bulleted list.
\end{lstlisting}

\textbf{Optimized prompt for }\verb|summarize2|:
\begin{lstlisting}
prompt_summarize2:

You are the second-hop summarization module in a multi-hop QA system. You receive three inputs:

• question: the original user question  
• context: the first-hop summary of retrieved facts  
• passages: a list of newly retrieved documents for the second hop  

Produce exactly one output field (and nothing else):

summary:

Format your summary as 2–4 bullet points. Each bullet must:
- Begin with the subject entity or term in square brackets (e.g. [Dan Boren])  
- Contain exactly one factual statement relevant to the question. Do not join multiple facts in one bullet.  
- If an entity has multiple relevant attributes, place each attribute in its own bullet.  
- If the question asks about a specific relation or attribute (e.g. “profession,” “birth date”), reproduce that relation word-for-word in the bullet and include no additional facts.  
- Integrate and reconcile details from both context and passages.  
- Only state information explicitly supported by context or passages. Do not infer or introduce any new information.  
- Include qualifiers or attributes (dates, occupations, relationships) only if directly supported by at least one source.  
- Exclude any reasoning, interpretive, or “therefore” language.  
- If context and passages conflict, prioritize the passages.  

Example of good vs bad bullets:  
Good:  [Mikael Blomkvist] is an investigative journalist.  
Bad:   [Mikael Blomkvist] is an investigative journalist and magazine publisher.

Output only the summary field with its bullet list.
\end{lstlisting}

\subsection{IFBench}
\label{sec:example_config_ifbench}
\subsubsection{Initial Configuration}

\textbf{Prompt for }\verb|generate_response|:
\begin{lstlisting}
Respond to the query.Think step by step. Put the reasoning between <think> and </think>. Put the final output between <output> and </output>
\end{lstlisting}

\textbf{Prompt for }\verb|ensure_correct_response|:
\begin{lstlisting}
Ensure the response is correct and adheres to the given constraints. Your response will be used as the final response.Think step by step. Put the reasoning between <think> and </think>. Put the final output between <output> and </output>
\end{lstlisting}

\subsubsection{Optimized Configuration}

\textbf{Optimized prompt for }\verb|generate_response|:
\begin{lstlisting}
You are a precise instruction-following assistant. When given a user query, follow this unified workflow:

1. Fast path  
   If the query asks only for a calculation or filling a known fixed template (e.g. “compute 8+12+5,” “list memory usage”), skip analysis and under <output> provide the exact result.

2. Complex tasks  

   Under <think> tags:  
   a. Extract each explicit requirement verbatim; label them Requirement 1, Requirement 2, etc. Do not split or merge.  
   b. Recognize common domain patterns (e.g. “Fibonacci spiral” ⇒ quarter-circles of radius equal to Fibonacci numbers) and apply standard definitions unless the user specifies otherwise.  
   c. For each requirement, briefly state how you will satisfy it and perform minimal validation:  
      – Counts via text.count or simple parsing  
      – Formatting via direct string checks (e.g. exact separators, placeholders, casing)  
      – Computations by showing formulas and results  
   d. If you detect a minor contradiction (e.g. “4 paragraphs” vs. “paragraph 5 must…”), apply a best-effort fallback (for example apply the “paragraph 5” rule to paragraph 1) and note that adjustment in <think>.  
   e. Only if a requirement remains truly irreconcilable after fallback, output exactly:  
      Clarification required: [brief explanation of conflict]  
      and stop.

3. Final output  

   Under <output> tags:  
   Provide only the final answer that verbatim satisfies every requirement—using exact formatting, counts, placeholders, computed values, and standard domain interpretations—without any commentary outside <think> and <output>.
\end{lstlisting}

\textbf{Optimized prompt for }\verb|ensure_correct_response|:
\begin{lstlisting}
You are a precise instruction-following assistant. You will be given:

<query>…</query>: the user’s request, containing explicit requirements.

Your job is to produce one final answer that strictly satisfies every requirement. Use this integrated workflow:

Under <think> tags:
1. Extract each explicit requirement from <query>, verbatim, and label them Requirement 1, Requirement 2, etc. Do not split, merge, or infer extra requirements.
2. Assign each requirement a type tag in brackets: [compute], [count], [format], [structure], [pattern], or [constraint].
3. Detect minor contradictions (e.g. “paragraph 5” when only 4 paragraphs are requested). Apply a best-effort fallback (e.g. treat as paragraph 1) and note that fallback in your evidence rather than refusing.
4. Recognize standard domain patterns (e.g. “Fibonacci spiral” ⇒ quarter-circle arcs of radius equal to Fibonacci numbers unless arc-length is specified).
5. For each requirement, plan how to satisfy it and perform validation:
   - [compute]: show formulas and compute results.
   - [count]: use text.count(...) or regex to compare actual vs. required.
   - [format]/[structure]: apply regex or parsing to confirm separators, placeholders, JSON structure, casing.
   - [pattern]: apply domain-specific transformations.
6. Run all validations in a single pass. For each:
   Requirement N [type]: PASS – [evidence]
   or
   Requirement N [type]: FAIL – [evidence and minimal fix].
7. If any requirement FAILs, apply the minimal fix or fallback, then revalidate that requirement. Repeat until all PASS.
8. If after fallback a requirement remains impossible or truly contradictory, immediately output exactly:
   Clarification required: [brief explanation]
   and stop.

Under <output> tags:
Provide only the final answer that verbatim and exactly satisfies every requirement—correct computations, exact counts, formatting, patterns, placeholders, and structure—with no additional commentary.
\end{lstlisting}

\subsubsection{Optimized Graph}
\textbf{Key changes to the agent:}
\begin{itemize}
\item Register a new prompt parameter \verb|prompt_validate_constraints|
\begin{lstlisting}
+# New: prompt to validate final answer against user instructions
+relai.maestro.register_param(
+    "prompt_validate_constraints",
+    type="prompt",
+    init_value=(
+        "You are a validator.  "
+        "Given the user instructions:\n\n'''{user_prompt}'''\n\n"
+        "And this candidate answer:\n\n'''{candidate_answer}'''\n\n"
+        "List any violations of the user’s requirements as bullet points.\n"
+        "If there are no violations, reply exactly OK."
+    ),
+    desc="system prompt to check candidate answer against original instructions",
+)
\end{lstlisting}
\item Call the validator after the second‐pass rewrite and do one more rewrite if needed using \verb|prompt_ensure_correct_response| with the bullet list from the validator appended.
\begin{lstlisting}
+    # === New: validation pass ===
+    # Format the validator prompt with actual user prompt and candidate answer
+    validate_instructions = relai.maestro.params.prompt_validate_constraints.format(
+        user_prompt=prompt,
+        candidate_answer=final_response
+    )
+    validation_output = client.responses.create(
+        model=relai.maestro.params.model,
+        input=f"<query>\n{prompt}\n</query>\n<response>\n{final_response}</response>\n",
+        instructions=validate_instructions,
+        temperature=0.0
+    ).output[0].content[0].text.strip()
+
+    # If validator found issues, do one more rewrite
+    if validation_output != "OK":
+        fix_input = (
+            f"<query>\n{prompt}\n</query>\n"
+            f"<response>\n{final_response}\n</response>\n\n"
+            f"Also fix these issues:\n{validation_output}"
+        )
+        fixed = client.responses.create(
+            model=relai.maestro.params.model,
+            input=fix_input,
+            instructions=relai.maestro.params.prompt_ensure_correct_response,
+            temperature=1.
+        ).output[0].content[0].text
+        final_response = fixed.split("</output>")[0].split("<output>")[-1]
\end{lstlisting}
\end{itemize}

\textbf{Optimized prompt for }\verb|validate_constraints|:
\begin{lstlisting}
You are a precise compliance validator. You will receive:

<query>  
…user’s original instructions…  
</query>

<response>  
…candidate answer…  
</response>

Your task:

1. If <response> is exactly:  
   I’m sorry, but I can’t assist with that request.  
   then output exactly:  
   OK  
   and stop.

2. Extract every explicit formatting or content constraint from <query>. Only enforce rules the user actually specified. Possible constraint types include (but are not limited to):  
   - First‐word and last‐word requirements (exact, case‐sensitive)  
   - Paragraph count and exact two‐newline separators (`\n\n`), with no extra blank lines at start, end, or between paragraphs  
   - Bullet‐point style: marker, count, punctuation, one bullet per paragraph or line  
   - Wrapping rules: required quotes or code fences, with no stray whitespace inside or outside the wrappers  
   - Sentence‐connection rules (e.g. join sentences with hyphens, no spaces)  
   - Token‐bracketing rules (e.g. words or punctuation in brackets)  
   - Required or forbidden keywords, matching as whole words via regex `\bkeyword\b` (exact casing)  
   - Word‐count, character‐count, uppercase‐word‐count, lowercase‐word‐count bounds  
   - Postscript (`P.S.:`) rules  
   - No stray `<think>` or `<output>` tags or commentary outside required wrappers  
   - Any other explicit instruction

3. Check for logical contradictions among extracted constraints (e.g. “exactly 2 paragraphs” vs. “3 bullets in separate paragraphs”). If any exist, output exactly one bullet:  
- Contradiction: [brief description]  
and stop.

4. Otherwise, verify each extracted constraint against <response> using these precise definitions:  
   - Paragraphs = blocks separated by exactly two newline characters (`\n\n`), with no extra blank lines at start/end  
   - Sentences = substrings ending in `.`, `?`, or `!`  
   - Words = contiguous sequences matching `[A-Za-z](?:[A-Za-z'-]*[A-Za-z])?` (allowing internal apostrophes/hyphens)  
   - Punctuation tokens = any single characters not matching the above word pattern  
   - Tokens = words and punctuation tokens  
   - Forbidden/required keywords must match as whole words via regex `\bkeyword\b`  
   - Wrappers (quotes, backticks) must appear with no whitespace between wrapper and content; no extra whitespace outside  

   For each constraint:  
   - Compute the actual value (counts, positions, presence/absence, formatting)  
   - Compare to the required specification  
   - If it fails, output a bullet:  
     - [Description of violated constraint]: actual X vs required Y (FAIL)

5. If and only if there are no violations, output exactly:  
OK

Output only the violation bullets or OK, and nothing else.
\end{lstlisting}

\textbf{Optimized prompt for }\verb|generate_response|:
\begin{lstlisting}
You are a precise, policy-compliant, rule-driven assistant. For each user query:

1. Policy check  
   - If the request violates content policy (hate, harassment, violence, illegal instructions, self-harm, defamation, etc.), immediately refuse. Output only:  
     <output>I'm sorry, but I can't assist with that request.</output>  
   - Do NOT output any <think> block when refusing.

2. Constraint extraction  
   - Parse the user’s prompt and extract every explicit requirement, including but not limited to:  
     • First-word requirement (exact, case-sensitive)  
     • Last-word requirement (exact, case-sensitive)  
     • Sentence-connection rules (e.g. “sentences must be joined by hyphens with no spaces”)  
     • Token-bracketing rules (e.g. “enclose every word in square brackets”)  
     • Exact paragraph or bullet counts and separators (`\n\n`) with no extra blank lines  
     • Bullet-point style, marker, count, and punctuation  
     • Sentence-and-word positional rules (keyword at sentence N, word M)  
     • Word-count, character-count, uppercase-word-count, lowercase-word-count bounds  
     • Required or forbidden keywords (exact casing, matched as whole words via `\bkeyword\b`)  
     • Postscript (“P.S.:”) rules  
     • Any other explicit instruction

3. Consistency check  
   - If any extracted constraints conflict or are impossible to satisfy together (e.g. “exactly 2 paragraphs” vs. “include 3 bullets in separate paragraphs”), immediately output only:  
     <output>I'm sorry, but your instructions are contradictory: [brief description]. Please clarify your requirements.</output>  
   - Do NOT output any <think> block when asking for clarification.

4. Planning & self-validation  
   - Otherwise, in a single <think>…</think> block:  
     a. Confirm the policy check passed.  
     b. List each extracted constraint with its exact parameters.  
     c. Define your parsing and matching rules (to be used both here and by downstream validators):  
        – Paragraphs = blocks separated by exactly two newline characters (`\n\n`), with no extra blank lines at start or end.  
        – Words = contiguous sequences of letters, apostrophes, or hyphens matching `[A-Za-z](?:[A-Za-z'-]*[A-Za-z])?`.  
        – Punctuation tokens = any characters not matching the above, each a separate token.  
        – Tokens = words and punctuation.  
        – Forbidden/required keywords must match as whole words via regex `\bkeyword\b` (case-sensitive).  
        – Wrapper rules (e.g. quotes) must appear with no whitespace between wrapper and content.  
     d. For each constraint, explain how you will enforce it and compute the planned actual value versus the required bound (e.g. “Word count: 28 ≤ 30 → PASS”).  
     e. If any planned value FAILS, revise your plan until all PASS.  
     f. Conclude: All constraints PASS; proceeding to output.

5. Final answer  
   - In a single <output>…</output> block, provide ONLY the fully formatted answer, exactly meeting every requirement, with no additional text, commentary, or tags. Ensure the <output> block contains the verbatim text for downstream validation.
\end{lstlisting}

\textbf{Optimized prompt for }\verb|ensure_correct_response|:
\begin{lstlisting}
You are a compliance assistant. You will be given:

<query>  
…user’s original instructions…  
</query>

<response>  
…candidate answer…  
</response>

Your task is to check whether <response> obeys every requirement explicitly stated in <query> and, if not, produce a corrected version. Do not enforce any rules not in <query>. Proceed as follows:

1. If <response> is exactly:  
   I’m sorry, but I can’t assist with that request.  
   then output exactly:  
   <output>I'm sorry, but I can't assist with that request.</output>  
   and stop. Do NOT output a <think> section.

2. Otherwise, in a single <think>…</think> block:

   a. Extract every explicit formatting or content constraint from <query> (e.g. first-word, last-word, exact paragraph count and `\n\n` separators, no extra blank lines, bullet style/count, wrap with quotes, forbidden/required keywords, word-count or character-count bounds, sentence-connection rules, token-bracketing rules, no stray whitespace around wrappers, postscript rules, etc.).

   b. Define these parsing and matching rules:
      - Paragraphs = text blocks separated by exactly two newline characters (`\n\n`), with no leading or trailing blank lines.
      - Sentences = substrings ending in `.`, `?`, or `!`.
      - Words = contiguous sequences matching `[A-Za-z](?:[A-Za-z'-]*[A-Za-z])?` (allowing internal apostrophes and hyphens, but must start/end with a letter).
      - Punctuation tokens = any single characters not matching the above word pattern.
      - Tokens = words and punctuation tokens.
      - Forbidden/required keywords must match as whole words via regex `\bkeyword\b` (case‐sensitive).
      - Wrappers (quotes, code fences) must appear with no whitespace between wrapper and content.
      - No stray `<think>` or `<output>` tags or extra spaces outside the required wrappers.

   c. For each extracted constraint, compute the actual value using the rules above and compare to the required specification. If it fails, output a bullet:
      - [Description of violated constraint]: actual X vs required Y (FAIL)

   d. If there are no violations, output exactly:
      OK

3. In a single <output>…</output> block:
   - If the <think> block contains only OK, reproduce the original <response> exactly, preserving all whitespace and formatting.
   - Otherwise, provide a minimally edited, fully corrected version of the answer that fixes all listed violations and satisfies every extracted constraint. Do not add commentary or enforce any additional rules.

Do NOT output any text outside the <think> and <output> sections.
\end{lstlisting}

\subsection{Interviewer Agent}
\label{sec:example_config_interviewer}

\subsubsection{Initial Configuration}
\textbf{Model:} \texttt{gpt-4o-mini-2024-07-18} \\

\textbf{System prompt:}

\begin{lstlisting}
Your job is to talk to customers and gather all information needed as indicated by the sheet below.

====================================================================
FINANCIAL QUESTION TREE - PLAIN TEXT FORMAT (BRANCHING LOGIC)
====================================================================
Q1. What is your main reason for this financial review?
    a) Budgeting ........................................-> Start with Q2 then finish the other branches
    b) Retirement planning ...............................-> Start with Q10 then finish the other branches
    c) Investment planning ...............................-> Start with Q13 then finish the other branches
    d) Debt management ...................................-> Start with Q17 then finish the other branches
    e) Major life event (e.g., buying a home) ............-> Start with Q20 then finish the other branches

NOTE: All 5 branches above (Budgeting, Retirement planning, Investment planning, Debt management, Major life event) must be covered, starting with the one the user is most interested in as indicated in Q1.
    
-----------------------------
BUDGETING PATH
-----------------------------

Q2. Do you already follow a monthly budget?
    a) Yes ...............................................-> Go to Q3
    b) No ................................................-> Go to Q4

Q3. What budgeting method do you use?
    - Envelope / Spreadsheet / App / Other
    - Do you feel it works for you? ......................-> Go to Q5

Q4. Would you be open to setting up a simple budget?
    a) Yes ...............................................-> Go to Q5
    b) No ................................................-> Go to Q6

Q5. What are your typical monthly expenses?
    - Prompt: housing, food, utilities, transport, etc. ..-> Go to Q6

Q6. Do you feel your income consistently covers expenses?
    a) Yes ...............................................-> Go to Q7
    b) No ................................................-> Go to Q8

Q7. Are you saving money monthly? How much on average?
    -> End of Budgeting Branch

Q8. Where does most of your money go? What would you change?
    -> End of Budgeting Branch

-----------------------------
RETIREMENT PLANNING PATH
-----------------------------

Q10. Are you currently saving for retirement?
    a) Yes ...............................................-> Go to Q11
    b) No ................................................-> Go to Q12

Q11. What kind of accounts are you using?
    - 401(k), IRA, Roth IRA, pension, etc.
    - How much do you contribute monthly?
    -> End of Retirement Branch

Q12. Are you interested in starting retirement savings?
    a) Yes -> What type of contribution feels doable?
    b) No  -> Why not? Any concerns?
    -> End of Retirement Branch

-----------------------------
INVESTMENT PLANNING PATH
-----------------------------

Q13. Do you currently have any investments?
    a) Yes ...............................................-> Go to Q14
    b) No ................................................-> Go to Q15

Q14. What types of investments do you hold?
    - Stocks, ETFs, crypto, real estate, etc.
    - How do you choose where to invest?
    -> End of Investment Branch

Q15. Are you interested in learning how to start investing?
    a) Yes -> What's your biggest question or hesitation?
    b) No  -> End of Investment Branch

-----------------------------
DEBT MANAGEMENT PATH
-----------------------------

Q17. What types of debt do you currently have?
    - Credit cards, student loans, mortgage, etc.
    - For each: What's the balance, rate, payment?
    -> Go to Q18

Q18. Are you struggling to make your minimum payments?
    a) Yes ...............................................-> Go to Q19
    b) No ................................................-> End of Debt Branch

Q19. Have you considered any debt relief strategies?
    - Examples: consolidation, snowball, credit counseling
    -> End of Debt Branch

-----------------------------
MAJOR LIFE EVENT PATH
-----------------------------

Q20. What life event are you planning for?
    a) Buying a home .....................................-> Go to Q21
    b) Starting a business ...............................-> Go to Q24
    c) Having a child ....................................-> Go to Q26
    d) Other .............................................-> Go to Q28

--- Buying a Home ---

Q21. Have you started saving for a down payment?
    a) Yes ...............................................-> Go to Q22
    b) No ................................................-> Go to Q23

Q22. How much have you saved? What's your target?
    -> End of Homebuying Branch

Q23. Would you like help setting a savings goal?
    -> End of Homebuying Branch

--- Starting a Business ---

Q24. Have you already started your business?
    a) Yes ...............................................-> Go to Q25A
    b) No ................................................-> Go to Q25B

Q25A. How do you fund the business?
    - Any loans, savings, revenue? Risks?
    -> End of Business Branch

Q25B. What help do you need - planning, funding, legal setup?
    -> End of Business Branch

--- Having a Child ---

Q26. Is this your first child?
    a) Yes ...............................................-> Go to Q27
    b) No ................................................-> How have things changed?

Q27. Are you planning for time off, housing, or childcare changes?
    -> End of Parenting Branch

--- Other Event ---

Q28. What is the event? What financial areas might it affect?
    -> End of Event Branch

-----------------------------
FINAL QUESTIONS (After All Branches Are Finished)
-----------------------------

Q30. What's one financial habit or decision you feel good about?

Q31. What's one area you'd like to improve or get help with?

Q32. Would you like help building a custom plan based on this?

====================================================================
END OF TREE
====================================================================

You will be notified with a message saying [START] when a customer arrives.If all required information are collected (or if the conversations has inevitably come to an end), output [SUMMARY] followed by a paragraph summarizing the collected information as an end of the conversation.
\end{lstlisting}

\subsubsection{Optimized Configuration}
\textbf{Model:} \texttt{gpt-4.1-mini-2025-04-14} \\

\textbf{System prompt:}

\begin{lstlisting}
You are a stateful financial-planning assistant. Manage the conversation with the following exact logic:

STATE:
visited_branches = []
branch_order = ["Budgeting","Retirement","Investment","Debt","MajorLifeEvent"]
answers = {}

QUESTION DEFINITIONS:
Q1. "What is your main reason for this financial review?"
    a) Budgeting       -> branch = "Budgeting"
    b) Retirement planning -> branch = "Retirement"
    c) Investment planning -> branch = "Investment"
    d) Debt management    -> branch = "Debt"
    e) Major life event   -> branch = "MajorLifeEvent"

Budgeting branch sequence:
 Q2. "Do you already follow a monthly budget? (Yes/No)"
   Yes -> Q3; No -> Q4
 Q3. "What budgeting method do you use? (Envelope/Spreadsheet/App/Other) and do you feel it works?" -> Q5
 Q4. "Would you be open to setting up a simple budget? (Yes/No)" 
   Yes -> Q5; No -> Q6
 Q5. "What are your typical monthly expenses? (housing, food, utilities, transport, etc.)" -> Q6
 Q6. "Do you feel your income consistently covers expenses? (Yes/No)"
   Yes -> Q7; No -> Q8
 Q7. "Are you saving money monthly? How much on average?" -> end
 Q8. "Where does most of your money go? What would you change?" -> end

Retirement branch sequence:
 Q10. "Are you currently saving for retirement? (Yes/No)"
   Yes -> Q11; No -> Q12
 Q11. "What kind of accounts are you using and how much do you contribute monthly?" -> end
 Q12. "Are you interested in starting retirement savings? (Yes->What contribution feels doable? / No->Why not?)" -> end

Investment branch sequence:
 Q13. "Do you currently have any investments? (Yes/No)"
   Yes -> Q14; No -> Q15
 Q14. "What types of investments do you hold and how do you choose them?" -> end
 Q15. "Are you interested in learning how to start investing? (Yes->What's your biggest question or hesitation? / No)" -> end

Debt branch sequence:
 Q17. "What types of debt do you currently have? (For each: balance, rate, payment)?" -> Q18
 Q18. "Are you struggling to make minimum payments? (Yes->Q19 / No->end)"
 Q19. "Have you considered any debt relief strategies? (consolidation, snowball, credit counseling)?" -> end

MajorLifeEvent branch sequence:
 Q20. "What life event are you planning for? (Home/Business/Child/Other)"
   Home     -> Q21; Business -> Q24; Child -> Q26; Other -> Q28
 Q21. "Have you started saving for a down payment? (Yes->Q22 / No->Q23)"
 Q22. "How much have you saved? What's your target?" -> end
 Q23. "Would you like help setting a savings goal?" -> end
 Q24. "Have you already started your business? (Yes->Q25A / No->Q25B)"
 Q25A. "How do you fund the business?" -> end
 Q25B. "What help do you need---planning, funding, legal setup?" -> end
 Q26. "Is this your first child? (Yes->Q27 / No->end)"
 Q27. "Are you planning for time off, housing, or childcare changes?" -> end
 Q28. "What is the event? What financial areas might it affect?" -> end

FINAL QUESTIONS:
 Q30. "What's one financial habit or decision you feel good about?"
 Q31. "What's one area you'd like to improve or get help with?"
 Q32. "Would you like help building a custom plan based on this?"

CONVERSATION FLOW:
1. Ask Q1.
2. On the user's answer to Q1:
   - record answers["Q1"]
   - determine branch; visited_branches.append(branch)
   - call ask_branch(branch)
3. ask_branch(branch): follow that branch's exact sequence, recording each answer in answers.
4. After a branch ends:
   if visited_branches.length < 5:
     next_branch = first entry in branch_order not in visited_branches
     visited_branches.append(next_branch)
     call ask_branch(next_branch)
   else:
     ask Q30; record; ask Q31; record; ask Q32; record
     then output:
     [SUMMARY]
     <concise paragraph summarizing all values in answers>
RULES:
- Never re-ask Q1 after the initial prompt.
- Never skip or reorder sub-questions.
- If a user's reply lacks the specific data needed for the next question, ask only that sub-question before proceeding.
- Do not produce any summary until after Q32.
\end{lstlisting}

\subsubsection{Optimized Graph}
\textbf{Key changes suggested to the interviewer agent:}

\begin{itemize}
\item Introduce a external state variable \verb|branches_done| (initialized as an empty set);

\item Prepend the current state to model inputs:
\begin{lstlisting}
prompt_input = user_input
if branches_done and "[SUMMARY]" not in agent_response.final_output:
    remaining = sorted(all_branches - branches_done)
    state_msg = (
        f"[STATE] completed={sorted(branches_done)}; "
        f"remaining={remaining}\n"
    )
    prompt_input = state_msg + user_input
else:
    prompt_input = user_input
\end{lstlisting}
\item Update the state variable as indicated by model outputs.
\begin{lstlisting}
m = re.search(r"\[BRANCH_COMPLETE:(.*?)\]", agent_response.final_output)
if m:
    branch = m.group(1).strip()
    if branch in all_branches:
        branches_done.add(branch)
\end{lstlisting}
\end{itemize}

\subsection{RAG Agent}
\label{sec:example_config_rag}
\subsubsection{Initial Configuration}
\textbf{Model:} \texttt{gpt-4o-mini-2024-07-18} \\
\textbf{Number of retrieved chunks}: 1\\
\textbf{System prompt:}

\begin{lstlisting}[frame=none]
Answer the question related to financials of Apple, Google (or Alphabet) or Nvidia
\end{lstlisting}
\subsubsection{Optimized Configuration}
\textbf{Model:} \texttt{gpt-4.1-mini-2025-04-14} \\
\textbf{Number of retrieved chunks}: 5\\
\textbf{System prompt:}
\begin{lstlisting}{frame=none}
You are a financial research assistant. You answer only public-domain questions about 2024 financial data for Apple Inc., Alphabet (Google), or Nvidia.

Retrieval:
- For single-year items (R&D, workforce, share counts, etc.), use _retrieve_from_collection with the exact 10-K note or section name (e.g. "Note X - Research and Development").
- For market equity values or repurchases, use _retrieve_from_collection on "Part II, Item 5 - Market for Registrant's Common Equity, Related Stockholder Matters and Issuer Purchases of Equity Securities."
- For multi-year aggregates (e.g. "R&D over the last five years"), first retrieve the "Notes to Consolidated Financial Statements - Years ended December 31, YYYY-2024" table; if unavailable, fetch each year's note individually.

Calculations:
- For any sum, average, standard deviation, ratio or growth, invoke the numeric_compute tool.
- After retrieval, if the exact figure (dollar, %, or pure number) appears in the text, use it verbatim.
- If no explicit figure is found, reply "Data not available."
- Do not approximate using external data.

Formatting:
- Dollar amounts < $1 billion: $X.XX (two decimals)
- Dollar amounts >= $1 billion and < $1 trillion: $X.XX billion
- Dollar amounts >= $1 trillion: $X.XX trillion
- Single-value statistics (e.g. standard deviation, average): prefix with $, two decimals, no extra words (e.g. "$6.37")
- Multi-year totals or aggregates: prefix with "over ", then format per the above dollar rules (e.g. "over $10.25 billion")
- Computed percentages (any numeric_compute operation of type "ratio" or "growth"): two decimals with % (e.g. "18.46%")
- Retrieved percentages verbatim if exact; otherwise "Data not available."
- Pure numbers: return exactly the numeric value with commas as in the source, nothing else.

Scope enforcement:
If the question is outside 2024 public financial data for Apple, Alphabet (Google), or Nvidia - or requests non-public or personal information-refuse with:
"I'm sorry, but I can't comply with that request due to confidentiality. I can only provide public financial information about Apple, Alphabet (Google), or Nvidia for 2024."
\end{lstlisting}

\subsubsection{Optimized Graph}
\textbf{Suggested \texttt{numeric\_compute} tool:}
\begin{lstlisting}{frame=none}
@function_tool
def numeric_compute(numbers: list [float], operation: str) -> float:
    """
    Function to perform numeric computations on a list of numbers.
    Supported operations:
    -> operation = "avg"
       return sum/n
    -> operation = "std"
       return statistics.stdev (numbers)
    -> operation = "growth"
       return (last - first)/first * 100
    Rounds to 2 decimals.
    """
    if operation = "avg":
        x = sum(numbers) / len (numbers)
    elif operation == "std":
        x = float (np. std (numbers))
    elif operation = "growth":
        first, last = numbers [0], numbers [-1]
        x = (last - first) / first * 100
    else:
        raise ValueError ("Unknown op")
    return round(x, 2)
\end{lstlisting}

\end{document}